\documentclass[sigconf]{acmart}

\usepackage{algorithm, algorithmic}
\usepackage{multirow}
\usepackage{xcolor}
\usepackage{subfigure}

\newtheorem{theorem}{\textbf{Theorem}}
\DeclareMathOperator*{\argmin}{arg\,min}

\AtBeginDocument{%
  }

\copyrightyear{2023}
\acmYear{2023}
\setcopyright{acmlicensed}\acmConference[WWW '23]{Proceedings of the ACM Web Conference 2023}{April 30--May 04, 2023}{Austin, TX, USA}
\acmBooktitle{Proceedings of the ACM Web Conference 2023 (WWW '23), April 30--May 04, 2023, Austin, TX, USA}
\acmPrice{15.00}
\acmDOI{10.1145/3543507.3583369}
\acmISBN{978-1-4503-9416-1/23/04}

\begin{document}

\title{CgAT: Center-Guided Adversarial Training for Deep Hashing-Based Retrieval}




\author{Xunguang Wang\texorpdfstring{$^1$}{}, Yiqun Lin\texorpdfstring{$^{1}$}{}, Xiaomeng Li\texorpdfstring{$^{1,2,\ast}$}{}}
\affiliation{%
	\institution{\texorpdfstring{$^1$}{}The Hong Kong University of Science and Technology}
    \city{Hong Kong}
    \country{China}
}
\affiliation{%
  \institution{$^2$The Hong Kong University of Science and Technology Shenzhen Research Institute}
  \city{Shenzhen}
  \state{Guangdong}
  \country{China}
}
\email{%
    xunguangwang@gmail.com, lyq211003@gmail.com, eexmli@ust.hk
}
\thanks{$\ast$ Xiaomeng Li is the corresponding author}

\renewcommand{\shortauthors}{Wang, et al.}
\renewcommand{\authors}{Xunguang Wang, Yiqun Lin, Xiaomeng Li}

\begin{abstract}
  Deep hashing has been extensively utilized in massive image retrieval because of its efficiency and effectiveness. However, deep hashing models are vulnerable to adversarial examples, making it essential to develop adversarial defense methods for image retrieval. Existing solutions achieved limited defense performance because of using weak adversarial samples for training and lacking discriminative optimization objectives to learn robust features. In this paper, we present a min-max based Center-guided Adversarial Training, namely \textbf{CgAT}, to improve the robustness of deep hashing networks through worst adversarial examples. Our key idea is to formulate a hash code (dubbed \textbf{center code}) as a discriminative semantic representation of the original sample, which can be used to guide the generation of the powerful adversarial example and as an accurate optimization objective for adversarial training. Specifically, we first formulate the center code as a semantically-discriminative representative of the input image content, which preserves the semantic similarity with positive samples and dissimilarity with negative examples. We prove that a mathematical formula can calculate the center code immediately. After obtaining the center codes in each optimization iteration of the deep hashing network, they are adopted to guide the adversarial training process. On the one hand, CgAT generates the worst adversarial examples as augmented data by \textbf{maximizing} the Hamming distance between the hash codes of the adversarial examples and the center codes. On the other hand, CgAT learns to mitigate the effects of adversarial samples by \textbf{minimizing} the Hamming distance to the center codes. Extensive experiments on the benchmark datasets demonstrate the effectiveness of our adversarial training algorithm in defending against adversarial attacks for deep hashing-based retrieval. Compared with the current state-of-the-art defense method, we significantly improve the defense performance by an average of 18.61\%, 12.35\%, and 11.56\% on FLICKR-25K, NUS-WIDE, and MS-COCO, respectively. The code is available at \url{https://github.com/xunguangwang/CgAT}.
\end{abstract}



\maketitle

\section{Introduction}
With the exponential growth of visual data on the Internet, hashing \cite{wang2017survey} has attracted much attention in content-based image retrieval. By mapping high-dimensional data into compact hash codes in binary space, hashing shows remarkable advantages in low time and space complexity for retrieval. In particular, deep hashing \cite{xia2014supervised,lai2015simultaneous, zhu2016deep, li2016feature, liu2016deep, li2017deep, cao2017hashnet, jiang2017asymmetric, cao2018deep, su2018greedy, wang2020deep, tu2021partial, hoe2021one, doan2022one, zhang2022probabilistic} that learns nonlinear hash functions through deep neural networks (DNNs) \cite{lecun2015deep, lin2020fpconv} has gradually become the leading technology in large-scale image retrieval since it achieves better performance than shallow hashing.


\begin{figure*}[t]
	\centering
	\subfigure[]{
		\begin{minipage}[t]{0.9\textwidth}
		\centering
		\includegraphics[width=0.95\textwidth]{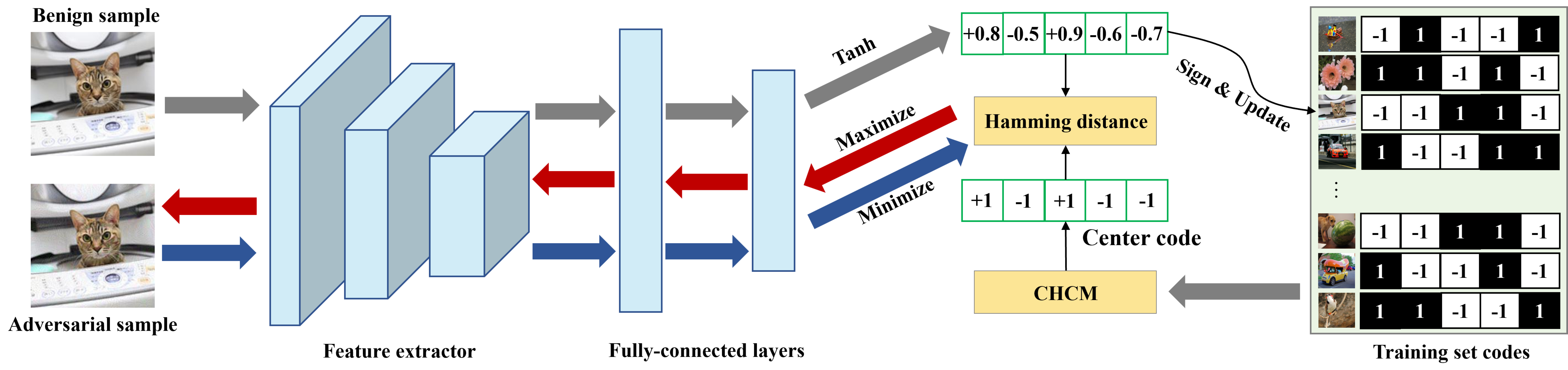}
		\end{minipage}
	}
	\subfigure[]{
		\begin{minipage}[t]{0.42\textwidth}
		\centering
		\includegraphics[width=1.0\linewidth]{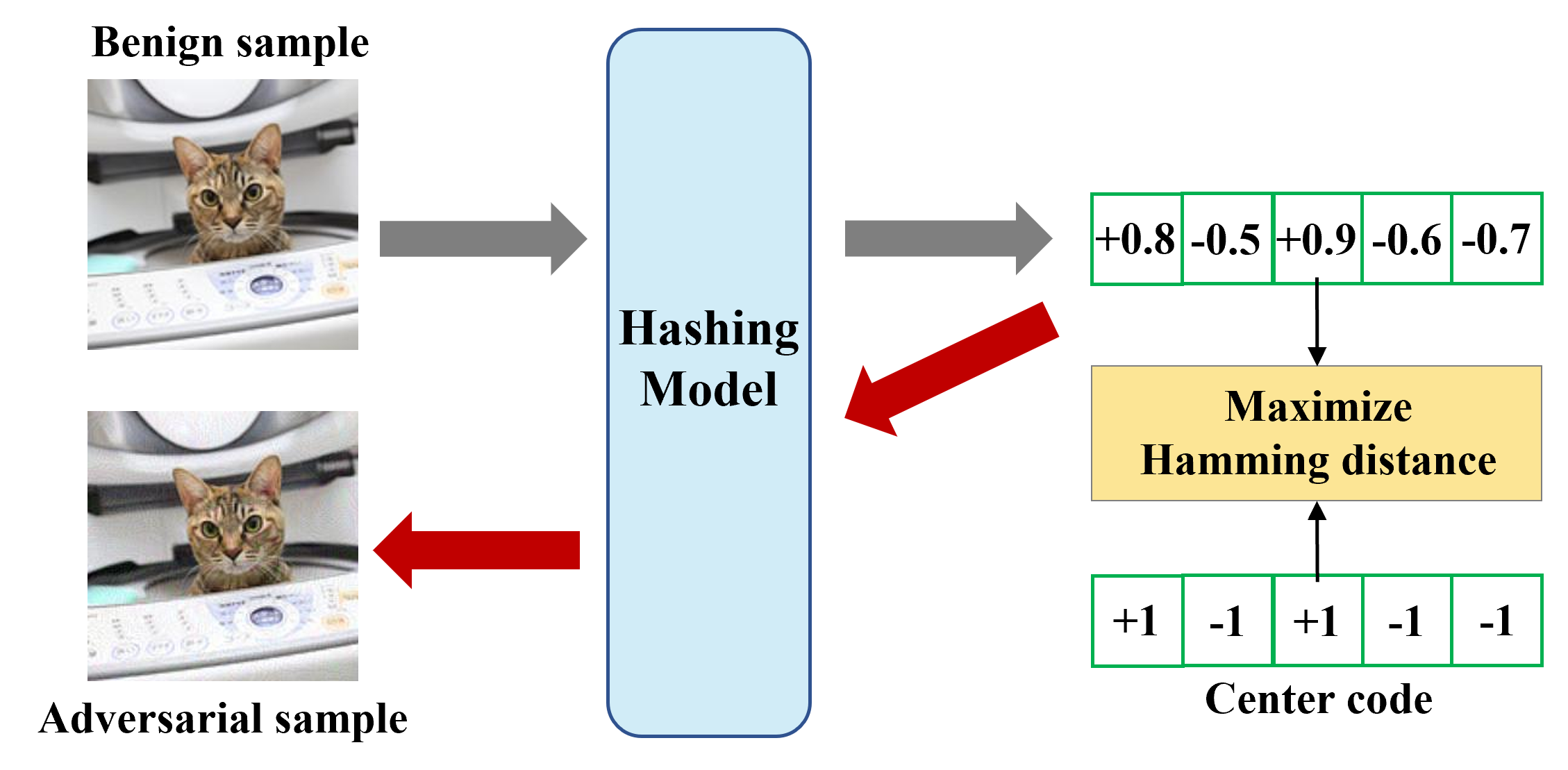}
		\end{minipage}
	}
	\hspace{6pt}
	\subfigure[]{
		\begin{minipage}[t]{0.42\textwidth}
		\centering
		\includegraphics[width=1.0\linewidth]{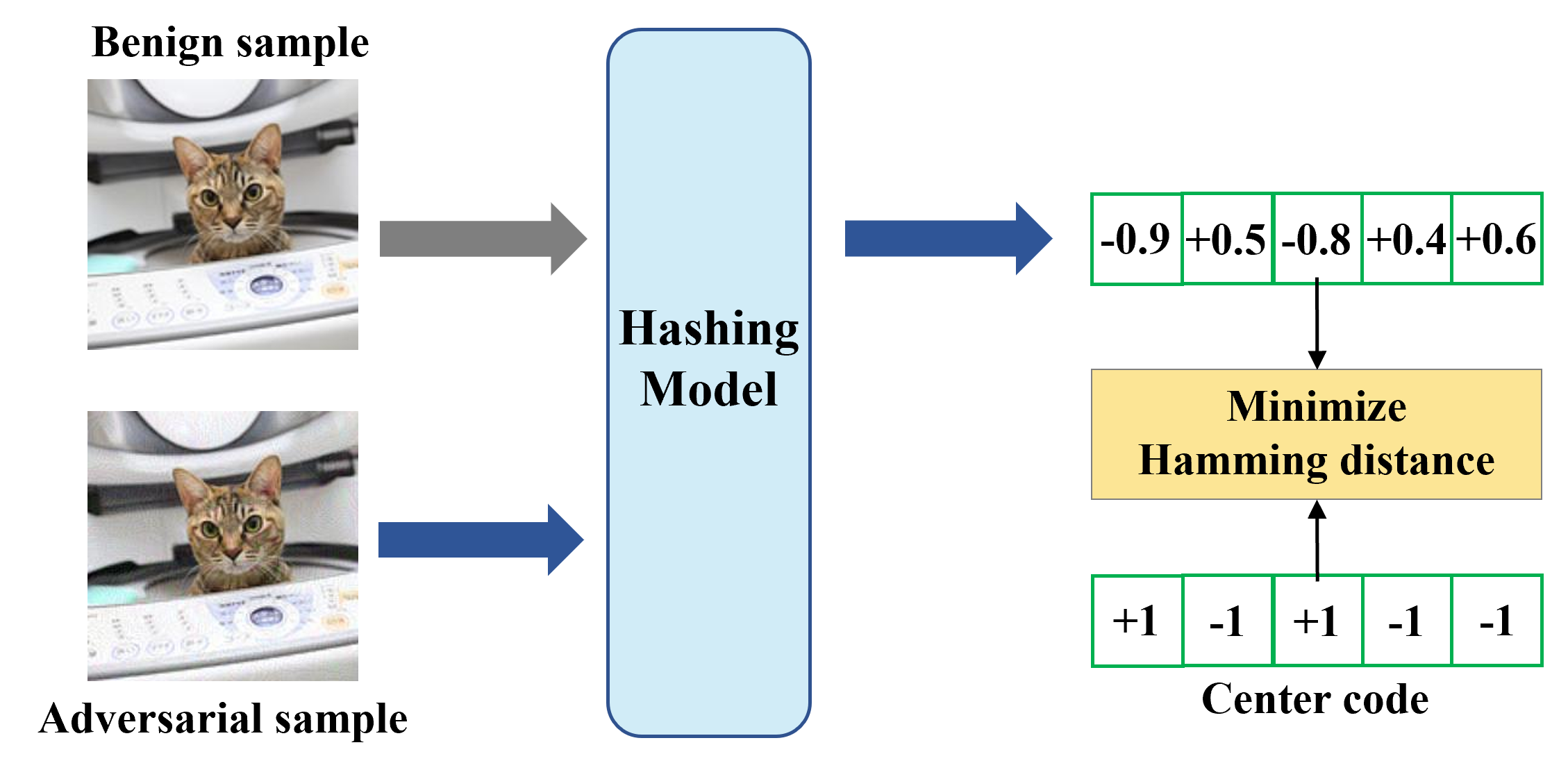}
		\end{minipage}
	}
	\centering
    \vspace{-2mm}
	\caption{The pipeline of the proposed adversarial training algorithm CgAT, where the \textcolor{gray}{gray}, \textcolor{red}{red}, and \textcolor{blue}{blue} arrows indicate forward, backward and forward propagation, respectively. The \textcolor{red}{red} arrow means constructing the adversarial sample with gradients. The \textcolor{blue}{blue} arrow represents inputting adversarial samples for adversarial training. (a) The overall framework of CgAT. Before each iteration of adversarial training, we calculate the center code with CHCM from the updated training set codes. Adversarial training of CgAT consists of alternating steps, \textit{i.e.}, (b) and (c). (b) Generating the adversarial example by maximizing the Hamming distance to the center code. (c) adversarial training by minimizing the distance of the adversarial sample to the center code.}
	\label{fig:framework}
	\vspace{-1mm}
\end{figure*}

Unfortunately, recent works~\cite{yang2018adversarial,bai2020targeted,wang2021prototype,wang2021targeted,zhang2021targeted,xiao2021you,lu2021smart} have revealed that deep hashing models are vulnerable to adversarial examples. Although these imperceptible samples are crafted by adding small perturbations to original samples, they are sufficient to fool deep hashing networks into making wrong predictions. 
Undoubtedly, such malicious attacks bring serious security threats to deep hashing-based image retrieval systems. For example, in a deep hashing-based face recognition system, adversarial examples can mislead the system into matching the faces of particular persons in the database, thereby successfully invading the system. Therefore, developing effective defense strategies in deep hashing-based retrieval is highly demanded.

Adversarial training~\cite{goodfellow2014explaining, madry2017towards} adopts the adversarial examples as training data to improve the adversarial robustness of DNNs and becomes the predominant defense solution in classification. However, it is difficult to directly transfer adversarial training from classification to hashing-based retrieval because popular deep hashing methods use semantic similarities between training samples as the optimization objective instead of categories. Wang \textit{et al.} \cite{wang2021targeted} proposed the first adversarial training method ATRDH in hashing-based retrieval. 
Notwithstanding adversarial training is demonstrated to be effective in deep hashing, some limitations still hinder the current defense performance. 
First, existing adversarial training methods in deep hashing-based retrieval have limited ability to generate strong adversarial samples which are beneficial to improve the defense capabilities of DNNs \cite{madry2017towards}.
Specifically, the targeted attack in ATRDH randomly selects target semantics to direct the generation of adversarial examples so that the maximized attack effects of adversarial samples can not be guaranteed.
Second, the objective function in ATRDH does not highlight the semantic intra-compactness and inter-separability between adversarial samples and original samples, which prevents adversarial training from learning discriminative features. 

To address the above shortcomings, this paper proposes Center-guided Adversarial Training (CgAT) for defending against adversarial attacks on deep hashing-based retrieval. The core idea is to design \textit{center code} to accurately represent the category-specific semantics of the input image. Then we adopt the center code to guide the generation of a strong adversarial sample. In addition, the center code is used as the optimization objective for adversarial training, as shown in Figure~\ref{fig:framework}.
Specifically, we design the {center code} as the discriminative representative of the image semantics, which simultaneously preserves the similarities to semantically relevant samples and the dissimilarities to irrelevant samples.
Benefiting the binary property of hash codes, we prove that the center code can be calculated directly through a simple mathematical formula using the proposed \textit{Continuous Hash Center Method} (CHCM).
The overall framework of CgAT is built on the center code. Firstly, CgAT obtains center codes of input data with CHCM on-the-fly during the adversarial training (precisely, before each optimization iteration of the adversarial training in CgAT).
Subsequently, CgAT conducts adversarial training with two alternating optimization steps. One is to construct severely biased adversarial samples by \textbf{maximizing} the Hamming distance between the hash codes of the adversarial examples and the center codes. Another is to optimize the hashing model with adversarial samples by \textbf{minimizing} the Hamming distance to the center codes. Due to the superiority of the dynamic center codes, our adversarial learning manner can significantly improve the ability of deep hashing networks to resist severe adversarial perturbations. In summary, our main contributions can be summarized as follows:
\begin{itemize}
    \item We design the dynamic center code as the precise semantic representative of the original image content for helping construct the adversarial training framework of deep hashing. It is noted that the center code can be obtained instantly by our proven mathematical formula in CHCM.
    \item A min-max adversarial training algorithm is provided, \textit{i.e.}, CgAT, to improve the adversarial robustness of deep hashing networks.
    \item Extensive experiments demonstrate that CgAT can be integrated into deep hashing frameworks and achieves state-of-the-art defense performance against adversarial attacks.
\end{itemize}

\section{Related Work}
\subsection{Deep Hashing based Image Retrieval}
Deep Hashing methods can be roughly categorized into unsupervised and supervised deep hashing. Unsupervised deep hashing methods \cite{salakhutdinov2009semantic,ghasedi2018unsupervised} learn hash functions from unlabeled training data by discovering their intrinsic semantic relations. For example, semantic hashing \cite{salakhutdinov2009semantic} adopts an auto-encoder to reconstruct input data and produces hash codes from hidden vectors.

Although the unsupervised schemes are more general, their retrieval performance is unsatisfactory due to the semantic gap dilemma \cite{smeulders2000content}. By contrast, supervised deep hashing methods use semantic labels or pairwise similarity scores as supervisory signals, yielding better retrieval precision than unsupervised ones.
As one of the pioneering deep hashing algorithms, CNNH \cite{xia2014supervised} adopts two independent steps for hash learning, \textit{i.e.}, designing target hash codes of training data and learning approximate hash function with DNN.
Recent deep hashing methods \cite{lai2015simultaneous, zhu2016deep, li2016feature, liu2016deep, li2017deep, cao2017hashnet, jiang2017asymmetric, cao2018deep, su2018greedy, wang2020deep, tu2021partial, hoe2021one, doan2022one, zhang2022probabilistic} focus on end-to-end learning schemes and pairwise similarities to improve the quality of hash codes. Li \textit{et al.} \cite{li2016feature} proposed a pairwise loss-based method to preserve the semantic similarity between data items in an end-to-end DNN, called Deep Pairwise-Supervised Hashing (DPSH). HashNet \cite{cao2017hashnet} prepends the pairwise similarity with the weight to alleviate data imbalance between positive and negative pairs.
ADSH \cite{jiang2017asymmetric} learns the pairwise similarity between the database and query set in an asymmetric way.
However, CSQ \cite{yuan2020central} improves the discriminability of generated hash codes by enforcing them close to the designed hash centers of labels. In addition, Doan \textit{et al.} proposed a kind of Sliced-Wasserstein-based distributional distance (named HSWD), which is dedicated to optimizing both coding balance and low-quantization error for high-quality hash codes.

\subsection{Adversarial Attack}
The \textit{adversarial attack} aims to deceive the DNN by constructing adversarial examples. Since Szegedy \textit{et al.} \cite{szegedy2013intriguing} and Biggio \textit{et al.} \cite{biggio2013evasion} discovered the intriguing properties of adversarial samples, various adversarial attack methods \cite{szegedy2013intriguing,goodfellow2014explaining,kurakin2016adversarial,moosavi2016deepfool,madry2017towards,carlini2017towards,dong2018boosting,papernot2017practical,chen2017zoo,ilyas2018black} have been proposed to fool well-trained classifiers. According to the permission to access the attacked model, adversarial attacks are divided into white-box attacks and black-box attacks.
Generally, white-box attacks have access to all the information of the model, so gradients are usually used to optimize adversarial samples. For example, FGSM \cite{goodfellow2014explaining} crafts adversarial samples by maximizing the loss along the gradient direction with a large step. As the multi-step variant of FGSM, I-FGSM \cite{kurakin2016adversarial} and PGD \cite{madry2017towards} iteratively update perturbations with small steps for better attack performance. In contrast, black-box attacks can only capture the outputs of the model. Therefore, the solution to the black-box attack is to estimate the gradients based on the output results \cite{papernot2017practical,chen2017zoo,ilyas2018black}. While the black-box attack is more challenging than the white-box, it is more practical in real-world scenarios.

In addition to image classification, it has been attracted surging interest to develop adversarial attacks for deep hashing based image retrieval \cite{yang2018adversarial,bai2020targeted,wang2021prototype,wang2021targeted,lu2021smart,zhang2021targeted}. Existing adversarial attack methods in deep hashing can be divided into two categories: \textit{\textbf{non-targeted attack}} and \textit{\textbf{targeted attack}}. For the non-targeted attack in hashing-based retrieval, it aims to fool the hashing model to retrieve results that are not related to the original image \cite{bai2020targeted,wang2021prototype}. Achieving the non-targeted attack by minimizing the hash code similarity between the adversarial example and the original sample, Yang \textit{et al.} \cite{yang2018adversarial} proposed {HAG}, the first adversarial attack method on deep hashing. Considering staying away from relevant images of the query, {SDHA} \cite{lu2021smart} generates more effective adversarial queries than HAG. As for the targeted attack, it aims that the retrieved images of the adversarial example are semantically relevant to the given target label \cite{bai2020targeted,wang2021prototype}. To reach the targeted attack, {P2P} and {DHTA} \cite{bai2020targeted} obtain the anchor code as the representative of the target label to direct the generation of the adversarial sample. Subsequently, Wang \textit{et al.} \cite{wang2021targeted} leverage a fully-connected network to learn the prototype code as target code for superior targeted attack, which is called {THA} in this paper. {ProS-GAN} \cite{wang2021prototype} designs an generative framework for efficient targeted hashing attack. Different from the above white-box scenarios, Xiao \textit{et al.} \cite{xiao2021you} proposed the targeted black-box attack {NAG} by enhancing the transferability of adversarial examples.

Similar to the center code, the prior works have representative codes of image semantics, \textit{i.e.}, anchor code \cite{bai2020targeted} and prototype code \cite{wang2021prototype, wang2021targeted}.
Different from the DHTA \cite{bai2020targeted}, whose anchor code is obtained by a few instances with the same semantics as the original sample, the proposed center code preserves the semantic similarity with relevant examples and dissimilarity with irrelevant samples from the training set. 
Moreover, we use the proven mathematical formula (\textit{i.e.}, CHCM) to instantly calculate the optimal center code on-the-fly in adversarial training, rather than learning the prototype codes \cite{wang2021prototype, wang2021targeted} through time-consuming neural networks as ProS-GAN and THA do. Hence, the prototype code is properly applied to attacking fixed hashing models, and our center code is better suited for alternatively optimized adversarial training.

\subsection{Adversarial Training}
Adversarial training \cite{goodfellow2014explaining, madry2017towards, pang2020bag} leverages the generated adversarial samples as training data to optimize the model for resistance to adversarial attacks, which is one of the most direct and effective ways to improve the robustness of neural networks in adversarial defense methods. The naive adversarial training \cite{goodfellow2014explaining} simultaneously optimizes the loss of clean and adversarial samples. Then Madry \textit{et al.} \cite{madry2017towards} reformulated adversarial training as a min-max problem and optimized DNNs with adversarial examples during training. As an effective regularization method, adversarial training is extensively used to improve the robustness and generalization of DNNs \cite{pang2020bag, utrera2020adversarially, li2021divergence}. 

In deep hashing, Wang \textit{et al.} \cite{wang2021targeted} proposed the first effective adversarial training algorithm based on the targeted attack (dubbed \textbf{ATRDH} here) by narrowing the semantic gap between the adversarial samples and the original samples in Hamming space. Compared to ATRDH, our CgAT minimizes the distance between the hash code of the adversarial example and the center code in a min-max framework, rather than directly reducing the similarity errors with original samples like ATRDH. Due to the semantic discrimination of center codes, our CgAT-trained hashing models gain better defense performances than ATRDH.

\section{Method}
\subsection{Preliminaries}
Suppose an attacked hashing model $F$ is learned from a training set $O=\{(\boldsymbol{x}_i,\boldsymbol{y}_i)\}_{i=1}^N$ that contains $N$ samples labeled with $C$ classes, where $\boldsymbol{x}_i$ indicates $i$-th image, and $\boldsymbol{y}_i=[y_{i1},y_{i2},...,y_{iC}]\in \{0,1\}^C$ denotes a label vector of $\boldsymbol{x}_i$. Notably, $y_{ij}=1$ means that $\boldsymbol{x}_i$ belongs to the $j$-th class. We use $\boldsymbol{B}=\{\boldsymbol{b}_i\}_{i=1}^N$ to describe the hash code set of $O$, where $\boldsymbol{b}_i$ is the hash code of $\boldsymbol{x}_i$.

\textbf{Deep Hashing Model.} For the given hashing model $F$, the hash code $\boldsymbol{b}_i$ of $\boldsymbol{x}_i$ is generated as follows:
\begin{equation}
    \begin{aligned}
        \boldsymbol{b}_i = {F}(\boldsymbol{x}_i) &= \operatorname{sign}(\boldsymbol{h}_i)= \operatorname{sign}(f_\theta(\boldsymbol{x}_i)), \\ \text{s.t.~} & \boldsymbol{b}_i \in \{-1,1\}^K ,
    \end{aligned}
\end{equation}
where $K$ denotes the hash code length, and $f(\cdot)$ with parameter $\theta$ is a DNN to
approximate hash function ${F}(\cdot)$. The final binary code $\boldsymbol{b}_i$ is obtained by applying the $\operatorname{sign}(\cdot)$ on the output $\boldsymbol{h}_i$ of $f_\theta(\boldsymbol{x}_i)$. Typically, we implement $f(\cdot)$ with a convolutional neural network (CNN). It is noted that we leverage the $\operatorname{tanh}$ activation at the end output layer of the CNN to simulate the sign function.


\subsection{Generation of Center Codes}
To construct the discriminative semantic representative (\textit{i.e.}, center code) of the image content for guiding the process of adversarial training, it is essential to preserve the semantic similarities from training data into the center codes. For credibility, this objective can be achieved by minimizing the Hamming distance between the center code of the original image and its semantically relevant samples, and simultaneously maximizing the distance from irrelevant samples. Thus, for the original sample $\boldsymbol{x}$, the objective of its center code $\boldsymbol{b}$ is defined as follows:
\begin{equation}
	\begin{aligned}
	    \min_{\boldsymbol{b}} \sum_{i}^{N_{\rm{p}}} w_i {D}(\boldsymbol{b}, {F}(\boldsymbol{x}_i^{(\rm{p})})) - \sum_{j}^{N_{\rm{n}}} w_j {D}(\boldsymbol{b}, {F}(\boldsymbol{x}_j^{(\rm{n})}))
	\end{aligned},
\label{eq:definition_center_code}
\end{equation}
where $F(\cdot)$ is the hashing function approximated by the deep model $f(\cdot)$, and $D(\cdot, \cdot)$ is a distance metric. $w_i$ and $w_j$ represent distance weights. $\boldsymbol{x}_i^{(\rm{p})}$ is a positive sample semantically related to the original sample, and $\boldsymbol{x}_i^{(\rm{n})}$ is a negative sample irrelevant to the original sample. Because the Eq. (\ref{eq:definition_center_code}) can push the center code close to hash codes of semantically related samples and away from those of unrelated samples, optimal semantic preserving in the center code would come true in theory. $N_{\rm{p}}$ and $N_{\rm{n}}$ are the number of the positive and the negative samples, respectively.

Specifically, Eq. (\ref{eq:definition_center_code}) in hashing is equivalent to finding $\boldsymbol{b}^\ast$ which is minimized the Hamming distance with the hash codes of positive samples and maximized the distance with those of negative instances, \textit{i.e.},
\begin{equation}
	\begin{aligned}
	    \boldsymbol{b}^{\ast}=\argmin_{\boldsymbol{b}\in\{-1,+1\}^K} \sum_{i}^{N_{\rm{p}}} w_i {D}_{\rm{H}}(\boldsymbol{b}, \boldsymbol{b}_i^{(\rm{p})}) - \sum_{j}^{N_{\rm{n}}} w_j {D}_{\rm{H}}(\boldsymbol{b}, \boldsymbol{b}_j^{(\rm{n})})
	\end{aligned},
\label{eq:obj_center_code}
\end{equation}
where $D_{\rm{H}}$ is the Hamming distance measure. $\boldsymbol{b}_i^{(\rm{p})}$ is the hash code of the positive sample $\boldsymbol{x}_i^{(\rm{p})}$, and $\boldsymbol{b}_j^{(\rm{n})}$ is the binary code of the negative sample $\boldsymbol{x}_j^{(\rm{n})}$. In other words, $\boldsymbol{b}_i^{(\rm{p})}$ and $\boldsymbol{b}_j^{(\rm{n})}$ are selected from $\boldsymbol{B}$.

Due to the binary characteristic of the hash code, we can directly calculate the optimal code (called \textit{center code} $\boldsymbol{b}^{\ast}$) in the problem (\ref{eq:obj_center_code}) by a simple mathematical formula, as shown in Theorem \ref{theo:chcm}.
\begin{theorem}
\label{theo:chcm}
Center code $\boldsymbol{b}^{\ast}$ formulated in Eq. (\ref{eq:obj_center_code}) can be calculated by the difference between the weighting sum of positive hash codes and that of negative ones, \textit{i.e.},
\begin{equation}
	\begin{aligned}
	    \boldsymbol{b}^{\ast} = \operatorname{sign}\left(\sum_{i}^{N_{\rm{p}}}w_{i}\boldsymbol{b}_i^{(\rm{p})} - \sum_{j}^{N_{\rm{n}}}w_{j}\boldsymbol{b}_j^{(\rm{n})} \right)
	\end{aligned}.
\label{eq:chcm}
\end{equation}
\end{theorem}
We name the way of obtaining the center code with Eq. (\ref{eq:chcm}) as Continuous Hash Center Method (CHCM). The proof of CHCM (\textit{i.e.}, Theorem \ref{theo:chcm}) is shown in Sec. \ref{sec:proof}.


In addition, we define the $w_i$ and $w_j$ as follows:
\begin{equation}
    w_{i/j} = \frac{N}{N_{\rm{p/n}}}\cdot s_{i/j},\quad N=N_{\rm{p}}+N_{\rm{n}}
\end{equation}
where $s_{i/j}$ denotes the similarity/dissimilarity between the center code $\boldsymbol{b}^{\ast}$ (\textit{i.e.}, the original image $\boldsymbol{x}$) and the $i/j$-th benign sample, which is usually determined by the hashing method. Assuming that $M$ represents the maximum similarity, and $z_{i/j}$ is the semantic similarity between the center code and $x_{i/j}$, then $s_i=z_i$, $s_j=M-z_j$. In this paper, all experiments are based on $M=1$ and $z_{i/j}$ ranges $[0,1]$. For instance, $z_i=1$ and $z_j=0$, so $s_i=1$ for similar pairs, and $s_j=1$ for dissimilar ones. $\frac{N}{N_{\rm{p/n}}}$ can balance the number difference between positive and negative samples.

\subsection{Center-guided Adversarial Training}
\textbf{Generating Adversarial Examples.}
In hashing-based retrieval, an adversarial example is to confuse the hashing model to retrieve irrelevant results. Since the center code represents the semantics of the image content, we can generate the adversarial sample by prompting its hash code away from the center code. Under the $L_{\infty}$ constraint, the generation of the adversarial sample is formulated as follows:
\begin{equation}
    \begin{aligned}
        \max_{\boldsymbol{x}^\prime} D_{\rm{H}} (\boldsymbol{b}^{\ast}, F(\boldsymbol{x}^\prime)), \quad
        \text{s.t.~} \|\boldsymbol{x} & - \boldsymbol{x}^\prime\|_{\infty} \leq \epsilon
    \end{aligned}.
    \label{eq:ca}
\end{equation}
For any hash code $\hat{\boldsymbol{b}}$ and $\check{\boldsymbol{b}}$, we have  $D_{\rm{H}}(\hat{\boldsymbol{b}}, \check{\boldsymbol{b}})=\frac{1}{2}(K-\hat{\boldsymbol{b}}^{\top}\check{\boldsymbol{b}})$. Accordingly, the Eq. (\ref{eq:ca}) is equivalent to:
\begin{equation}
    \begin{aligned}
        \max_{\boldsymbol{x}^\prime} = -\frac{1}{K}(\boldsymbol{b}^{\ast})^{\top} F(\boldsymbol{x}^\prime), \quad
        \text{s.t.~} \|\boldsymbol{x} & - \boldsymbol{x}^\prime\|_{\infty} \leq \epsilon
    \end{aligned}.
    \label{eq:att_t}
\end{equation}
To alleviate the vanishing gradient problem in $F$, we actually replace $F$ with $f_{\theta}$. Hence, the objective $\mathcal{L}_{ca}$ for generating adversarial examples is described as follows:
\begin{equation}
    \begin{aligned}
        \max_{\boldsymbol{x}^\prime} ~\mathcal{L}_{ca}= -\frac{1}{K}(\boldsymbol{b}^{\ast})^{\top} f_{\theta}(\boldsymbol{x}^\prime), \quad
        \text{s.t.~} \|\boldsymbol{x} & - \boldsymbol{x}^\prime\|_{\infty} \leq \epsilon
    \end{aligned}.
    \label{eq:att_p}
\end{equation}
Unlike HAG and SDHA using SGD \cite{robbins1951stochastic} or Adam \cite{kingma2014adam} optimizer \cite{yang2018adversarial,lu2021smart} with more than $1,500$ iterations, this paper adopts PGD \cite{madry2017towards} to optimize $\boldsymbol{x}^\prime$ with $T$ iterations for efficiency, \textit{i.e.},
\begin{equation}
    \begin{aligned}
        \boldsymbol{x}_T^\prime = {\mathcal{S}}_{\epsilon}(\boldsymbol{x}_{T-1}^\prime + \alpha \cdot \operatorname{sign}(\Delta_{\boldsymbol{x}_{T-1}^\prime}\mathcal{L}_{ca})),
		\quad \boldsymbol{x}^{\prime}_0 = \boldsymbol{x},
    \end{aligned}
\end{equation}
where $\alpha$ is the step size, and $\mathcal{S}_{\epsilon}$ project $\boldsymbol{x}^\prime$ into the $\epsilon$-ball \cite{madry2017towards} of $x$.

\begin{table}[t]
\caption{MAP (\%) of defense methods under multiple attacks for FLICKR-25K. Notably, CgAT and ATRDH \cite{wang2021targeted} are built on DPSH \cite{li2016feature}, \textit{i.e.}, $\mathcal{L}_{ori}$ in Eq. (\ref{eq:obj_cat}) for CgAT is the objective function of DPSH. "Clean" indicates we test MAP with original samples. P2P \cite{bai2020targeted}, DHTA \cite{bai2020targeted}, ProS-GAN \cite{wang2021prototype}, THA \cite{wang2021targeted}, HAG \cite{yang2018adversarial} and SDHA \cite{lu2021smart} are adversarial attack methods which evaluate the robustness of deep hashing models with generated adversarial examples.}
\resizebox{\columnwidth}{!}{
\begin{tabular}{c|c|c|cccccc}
\toprule
$K$              & Method & Clean & P2P & DHTA &ProS-GAN & THA & HAG & SDHA \\ \hline
\multirow{4}{*}{16 bits} & DPSH \cite{li2016feature}   &82.53  &41.00 &33.21 &59.87 &33.88 &27.36 &19.46 \\
                         & ATRDH \cite{wang2021targeted}  &73.48 &51.48 &49.52 &74.39 &52.59 &40.72 &39.56  \\
                         & CgAT (ours)   &74.88 &\textbf{59.75} &\textbf{56.31} &\textbf{74.55} &\textbf{59.55}  &\textbf{52.69}  &\textbf{50.38} \\
                     &$\uparrow$ &\color{orange}{1.4} &\color{orange}{8.27} &\color{orange}{7.59} &\color{orange}{0.39}  &\color{orange}{6.61} &\color{orange}{13.41} &\color{orange}{10.82} \\ \hline
\multirow{4}{*}{32 bits} & DPSH \cite{li2016feature}   &84.00 &40.26 &32.78 &71.81 &36.17 &26.72 &19.34 \\
                         & ATRDH \cite{wang2021targeted}  &76.65 &52.04 &47.25 &77.14 &50.67 &38.24 &39.18  \\
                         & CgAT (ours)   &77.12 &\textbf{58.07} &\textbf{53.68} &\textbf{77.51} &\textbf{57.09}  &\textbf{49.91}  &\textbf{53.42}  \\
                     &$\uparrow$ &\color{orange}{0.47} &\color{orange}{6.03} &\color{orange}{6.43} &\color{orange}{0.37} &\color{orange}{6.42} &\color{orange}{11.67} &\color{orange}{14.24} \\ \hline
\multirow{4}{*}{64 bits} & DPSH \cite{li2016feature}   &84.53 &40.78 &34.53 &79.85 &36.49 &27.43 &26.54  \\
                         & ATRDH \cite{wang2021targeted}  &77.40  &56.34 &52.57 &77.75 &54.85 &43.98 &39.14   \\
                         & CgAT (ours)   &79.50 &\textbf{59.79} &\textbf{54.58} &\textbf{79.21} &\textbf{56.88} &\textbf{50.85} &\textbf{69.91}     \\ 
                     &$\uparrow$ &\color{orange}{2.10} &\color{orange}{3.45} &\color{orange}{2.01} &\color{orange}{1.46} &\color{orange}{2.03} &\color{orange}{6.87} &\color{orange}{30.77} \\ \bottomrule
\end{tabular}
}
\label{tab:flickr-25k}
\end{table}

\textbf{Adversarial Training.}
After generating adversarial examples, we hope to use them as augmentation data to optimize the deep hashing model for defense, \textit{i.e.}, adversarial training. The most direct and effective idea of adversarial training for deep hashing is to reconstruct the semantic similarity between adversarial and original samples, which can ensure that the adversarial examples can still retrieve relevant results from database. Since the center code as the optimal code preserves the similarity with the original samples, we can increase the semantic similarities between the adversarial samples and the benign samples by minimizing the Hamming distance between the hash code of adversarial sample and the center code. Thus, we define the adversarial loss for defense as follows:
\begin{equation}
    \begin{aligned}
        \min \sum_{i=1 }^N D_{\rm{H}} (\boldsymbol{b}^{\ast}_{i}, F(\boldsymbol{x}_i^\prime))
    \end{aligned},
    \label{eq:adv_h}
\end{equation}
where $\boldsymbol{b}^{\ast}_{i}$ is the center code of the original image $\boldsymbol{x}_i$, and $\boldsymbol{x}_i^\prime$ is the adversarial sample of $\boldsymbol{x}_i$.
Due to $D_{\rm{H}}(\hat{\boldsymbol{b}}, \check{\boldsymbol{b}})=\frac{1}{2}(K-\hat{\boldsymbol{b}}^{\top}\check{\boldsymbol{b}})$, Eq. (\ref{eq:adv_h}) is equivalent to:
\begin{equation}
    \begin{aligned}
        \min \mathcal -\sum_{i=1 }^N\frac{1}{K}(\boldsymbol{b}^{\ast}_{i})^{\top}F(\boldsymbol{x}_i^\prime)
    \end{aligned}.
\end{equation}
Similarly, we use $f_\theta$ instead of $F$ due to the gradient vanishing problem of $F$. Thus, the adversarial loss $\mathcal{L}_{adv}$ for optimizing the hashing model with adversarial samples is formulated as follows:
\begin{equation}
    \begin{aligned}
        \min \mathcal{L}_{adv}=-\sum_{i=1 }^N\frac{1}{K}(\boldsymbol{b}^{\ast}_{i})^{\top}f_{\theta}(\boldsymbol{x}_i^\prime)
    \end{aligned}.
\end{equation}
In summary, we propose the center-guided adversarial training (CgAT), and its objective is described as
\begin{equation}
    \begin{aligned}
        \min_{\theta}\mathcal{L}_{cat} = \mathcal{L}_{ori} + \lambda\mathcal{L}_{adv}
    \end{aligned},
    \label{eq:obj_cat}
\end{equation}
where $\lambda$ is the weighting factor. $\mathcal{L}_{ori}$ is the loss function of the original hashing method, which can ensure the basic performance of the hashing network.

Notably, the proposed CgAT is an adversarial learning algorithm. Hence, the overall objective of CgAT can be written as a minimax form:
\begin{equation}
    \begin{aligned}
        \min_{\theta}[\max_{\boldsymbol{x}^\prime}\mathcal{L}_{cat}(\boldsymbol{x}^\prime,y;\theta)]
    \end{aligned}.
\end{equation}
Like other adversarial learning algorithms, the proposed CgAT adopts an alternate strategy to retrain hashing models, as outlined in Algorithm \ref{alg:CgAT}. Firstly, we select hash codes of positive and negative samples for training images from $\boldsymbol{B}$ and obtain the center codes of input data by CHCM. Then, fixing the parameter $\theta$ of the hashing network, we generate adversarial examples with PGD. Subsequently, we use original and adversarial samples to optimize $\theta$ over Eq. (\ref{eq:obj_cat}). Finally, $\boldsymbol{B}$ is updated with input samples. Due to updating $\boldsymbol{B}$ constantly, the center codes based on it are \textbf{dynamic}.

\renewcommand{\algorithmicrequire}{\textbf{Input:}}
\renewcommand{\algorithmicensure}{\textbf{Output:}}
\begin{algorithm}[t]
	\caption{Center-guided Adversarial Training}
	\label{alg:CgAT}
	\begin{algorithmic}[1]
		\REQUIRE
		a training dataset $O=\{(\boldsymbol{x}_i, \boldsymbol{y}_i)\}_{i=1}^N$, pre-trained hashing model $F(\cdot)=\operatorname{sign}(f_\theta(\cdot))$, hash code set $\boldsymbol{B}$ of the training set $O$, training epochs $E$, batch size $n$, learning rate $\eta$, step size $\alpha$, perturbation budget $\epsilon$, attack iterations $T$, weighting factor $\lambda$
		\STATE \textbf{Initialize:} $\boldsymbol{B}=\{F(\boldsymbol{x}_i)\}_{i=1}^N$
		\FOR{$iter=1...E$}
		\FOR{image batch $\{x_i,y_i\}_{i=1}^n$}
		\STATE for each $x_i$, calculate its center code $\boldsymbol{b}^{\ast}_{i}$ with CHCM according to $\boldsymbol{B}$, \textit{i.e.}, Eq. (\ref{eq:chcm})
		\STATE optimize the adversarial samples by PGD with $T$ iterations: \\
		\quad $\boldsymbol{x}_i^\prime \gets\mathcal{S}_{\epsilon}(\boldsymbol{x}_i^\prime+\alpha\cdot \operatorname{sign}(\Delta_{\boldsymbol{x}_i^\prime}\mathcal{L}_{cat}(\boldsymbol{x}_i^\prime, \boldsymbol{b}^{\ast}_{i})))$,~$\forall~i=1,...,n$
		\STATE update $\theta$ with the gradient descent:\\
		\quad $\theta \gets \theta-\eta\Delta_{\theta} \frac{1}{n}\sum_{i=1}^n\mathcal{L}_{cat}(\boldsymbol{x}_i, \boldsymbol{x}_i^\prime, \boldsymbol{b}^{\ast}_{i};\theta)$
		\STATE update $\boldsymbol{B}$: $\boldsymbol{B}_i=F(\boldsymbol{x}_i)$, ~$\forall~i=1,...,n$
		\ENDFOR
		\ENDFOR
		\ENSURE network parameter $\theta$
	\end{algorithmic}
\end{algorithm}

\begin{table*}[t]
\centering
\begin{minipage}[t]{\columnwidth}
\centering
\caption{MAP (\%) of defense methods under various attacks for NUS-WIDE. Notably, CgAT and ATRDH \cite{wang2021targeted} are built on DPSH \cite{li2016feature} (The $\mathcal{L}_{ori}$ in Eq. (\ref{eq:obj_cat}) for CgAT is the objective function of DPSH).}
\vspace{-0.3cm}
\resizebox{\columnwidth}{!}{
\begin{tabular}{c|c|c|cccccc}
\toprule
$K$              & Method & Clean & P2P & DHTA &ProS-GAN & THA & HAG & SDHA \\ \hline
\multirow{4}{*}{16 bits} & DPSH \cite{li2016feature}   &80.79  &29.11 &23.34 &24.36 &22.04 &15.78 &15.26 \\ 
                         & ATRDH \cite{wang2021targeted}  &66.79 &56.14 &55.74 &66.81 &56.02 &48.98 &49.56  \\
                         & CgAT (ours)   &67.15 &\textbf{63.70} &\textbf{63.33} &\textbf{67.20} &\textbf{62.63}  &\textbf{62.39}  &\textbf{60.65} \\
                     &$\uparrow$ &\color{orange}{0.36} &\color{orange}{7.56} &\color{orange}{7.59} &\color{orange}{0.39}  &\color{orange}{6.61} &\color{orange}{13.41} &\color{orange}{11.09} \\ \hline
\multirow{4}{*}{32 bits} & DPSH \cite{li2016feature}   &82.20    &29.50 &23.38 &29.84 &21.79 &17.18 &17.82 \\
                         & ATRDH \cite{wang2021targeted}  &70.38 &48.31 &47.96 &69.64 &51.14 &47.17 &46.14  \\
                         & CgAT (ours)   &72.05 &\textbf{64.81} &\textbf{64.56} &\textbf{71.07} &\textbf{63.02}  &\textbf{64.62}  &\textbf{64.22}  \\
                     &$\uparrow$ &\color{orange}{1.67} &\color{orange}{16.50} &\color{orange}{16.60} &\color{orange}{1.43} &\color{orange}{11.88} &\color{orange}{17.45} &\color{orange}{18.08} \\ \hline
\multirow{4}{*}{64 bits} & DPSH \cite{li2016feature}   &82.24 & 29.44    &20.48 &51.18 &22.62 &17.88 &25.33 \\
                         & ATRDH \cite{wang2021targeted}  &72.31  &52.96 &51.72 &72.42 &54.81 &45.49 &58.24   \\
                         & CgAT (ours)   &77.38 &\textbf{60.61} &\textbf{58.61} &\textbf{77.55} &\textbf{61.39} &\textbf{62.65} &\textbf{66.13}     \\ 
                     &$\uparrow$ &\color{orange}{5.07} &\color{orange}{7.65} &\color{orange}{6.89} &\color{orange}{5.13} &\color{orange}{6.58} &\color{orange}{17.16} &\color{orange}{7.89} \\ \bottomrule
\end{tabular}
}
\label{tab:nus-wide}
\end{minipage}
\hfill
\begin{minipage}[t]{\columnwidth}
\centering
\caption{MAP (\%) of defense methods under attacks for MS-COCO. Notably, CgAT and ATRDH \cite{wang2021targeted} are built on DPSH \cite{li2016feature}, \textit{e.g.}, we replace $\mathcal{L}_{ori}$ in Eq. (\ref{eq:obj_cat}) with the objective function of DPSH.}
\vspace{-0.3cm}
\resizebox{\columnwidth}{!}{
\begin{tabular}{c|c|c|cccccc}
\toprule
$K$              & Method & Clean & P2P & DHTA &ProS-GAN & THA & HAG & SDHA \\ \hline
\multirow{4}{*}{16 bits} & DPSH \cite{li2016feature}   &60.10  &20.68 &17.02 &30.65 &25.27 &10.14 &10.69 \\
                         & ATRDH \cite{wang2021targeted}  &50.13 &35.06 &34.63 &48.54 &31.90 &29.14 &29.29  \\
                         & CgAT (ours)   &53.44 &\textbf{45.23} &\textbf{45.31} &\textbf{50.01} &\textbf{46.56}  &\textbf{42.15}  &\textbf{38.05} \\
                     &$\uparrow$ &\color{orange}{3.31} &\color{orange}{10.17} &\color{orange}{10.68} &\color{orange}{1.47} &\color{orange}{14.66} &\color{orange}{13.01} &\color{orange}{8.76} \\ \hline
\multirow{4}{*}{32 bits} & DPSH \cite{li2016feature}   &60.50 &25.99 &21.34 &44.06 &27.41 &9.22 &10.07 \\
                         & ATRDH \cite{wang2021targeted}  &48.04 &36.65 &35.74 &48.00 &33.60 &29.18 &29.47  \\
                         & CgAT (ours)   &54.13 &\textbf{44.51} &\textbf{44.15} &\textbf{53.97} &\textbf{51.08}  &\textbf{47.40}  &\textbf{40.37}  \\
                     &$\uparrow$ &\color{orange}{6.09} &\color{orange}{7.86} &\color{orange}{8.41} &\color{orange}{5.97} &\color{orange}{17.48} &\color{orange}{18.22} &\color{orange}{10.90} \\ \hline
\multirow{4}{*}{64 bits} & DPSH \cite{li2016feature}  &64.19 & 22.37 &18.70 &51.26 &30.86 &9.68 &10.06 \\
                         & ATRDH \cite{wang2021targeted}  &48.90  &35.85 &35.51 &48.89 &33.64 &28.47 &29.98   \\
                         & CgAT (ours)   &57.16 &\textbf{43.93} &\textbf{42.81} &\textbf{51.97} &\textbf{40.81} &\textbf{42.77} &\textbf{45.00}     \\ 
                     &$\uparrow$ &\color{orange}{8.26} &\color{orange}{8.08} &\color{orange}{7.30} &\color{orange}{3.08} &\color{orange}{7.17} &\color{orange}{14.30} &\color{orange}{15.02} \\ \bottomrule
\end{tabular}
}
\label{tab:ms-coco}
\end{minipage}
\end{table*}

\begin{figure*}[ht]
\vspace{-0.6cm}
\centering
\subfigure[FLICKR-25K]{ 
    \includegraphics[width=0.31\textwidth]{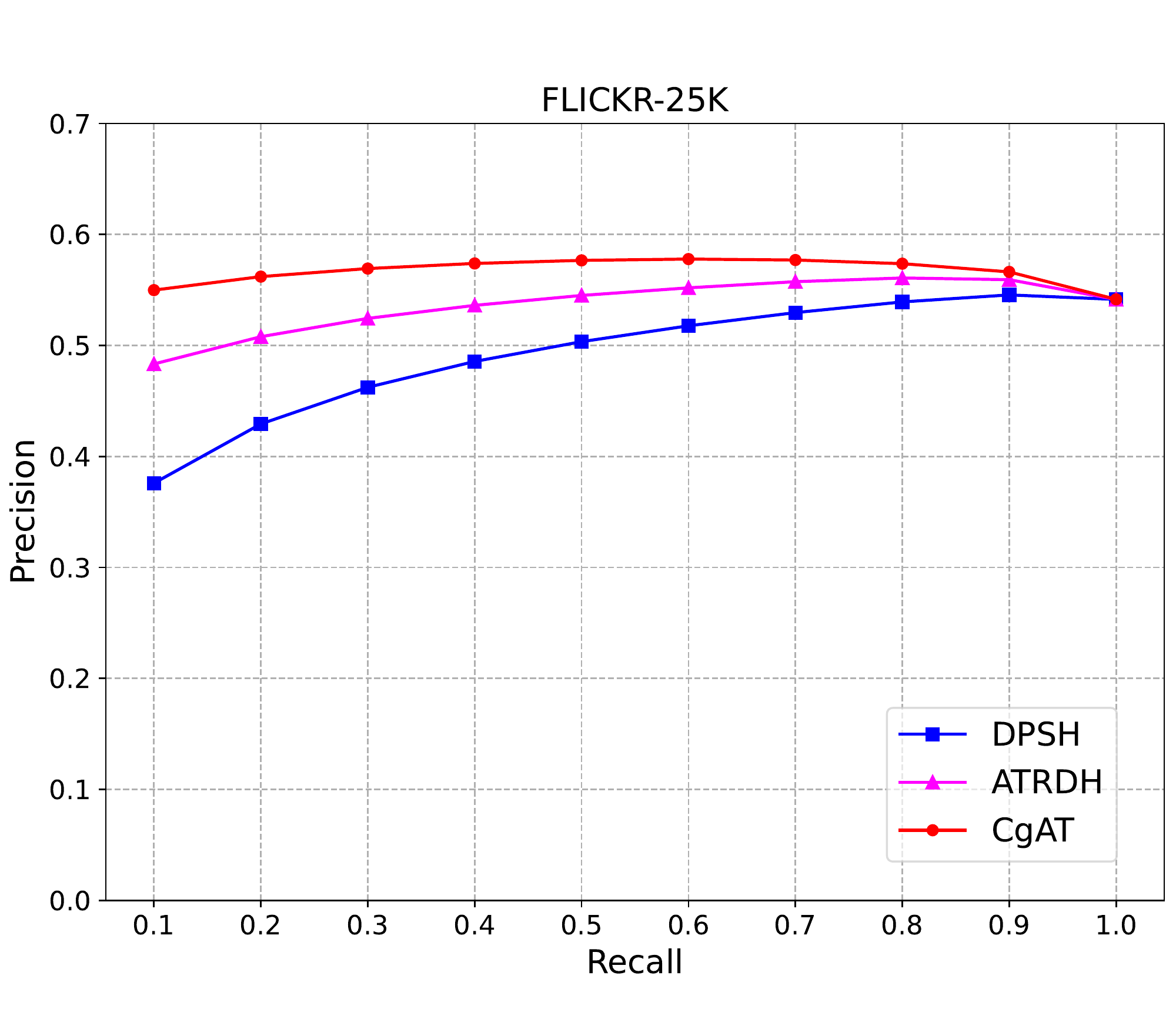}}
\subfigure[NUS-WIDE]{
    \includegraphics[width=0.31\textwidth]{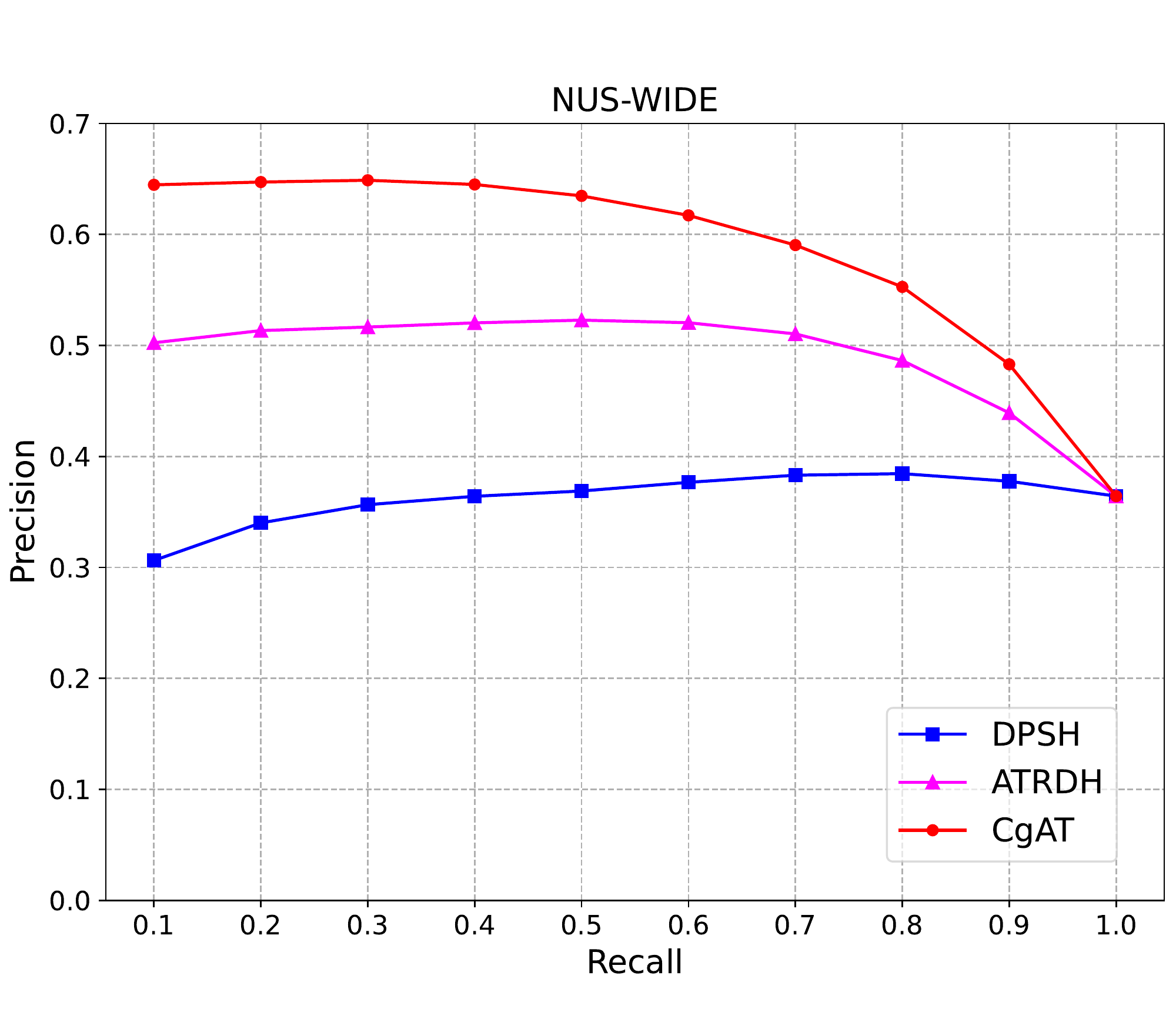}}
\subfigure[MS-COCO]{
    \includegraphics[width=0.31\textwidth]{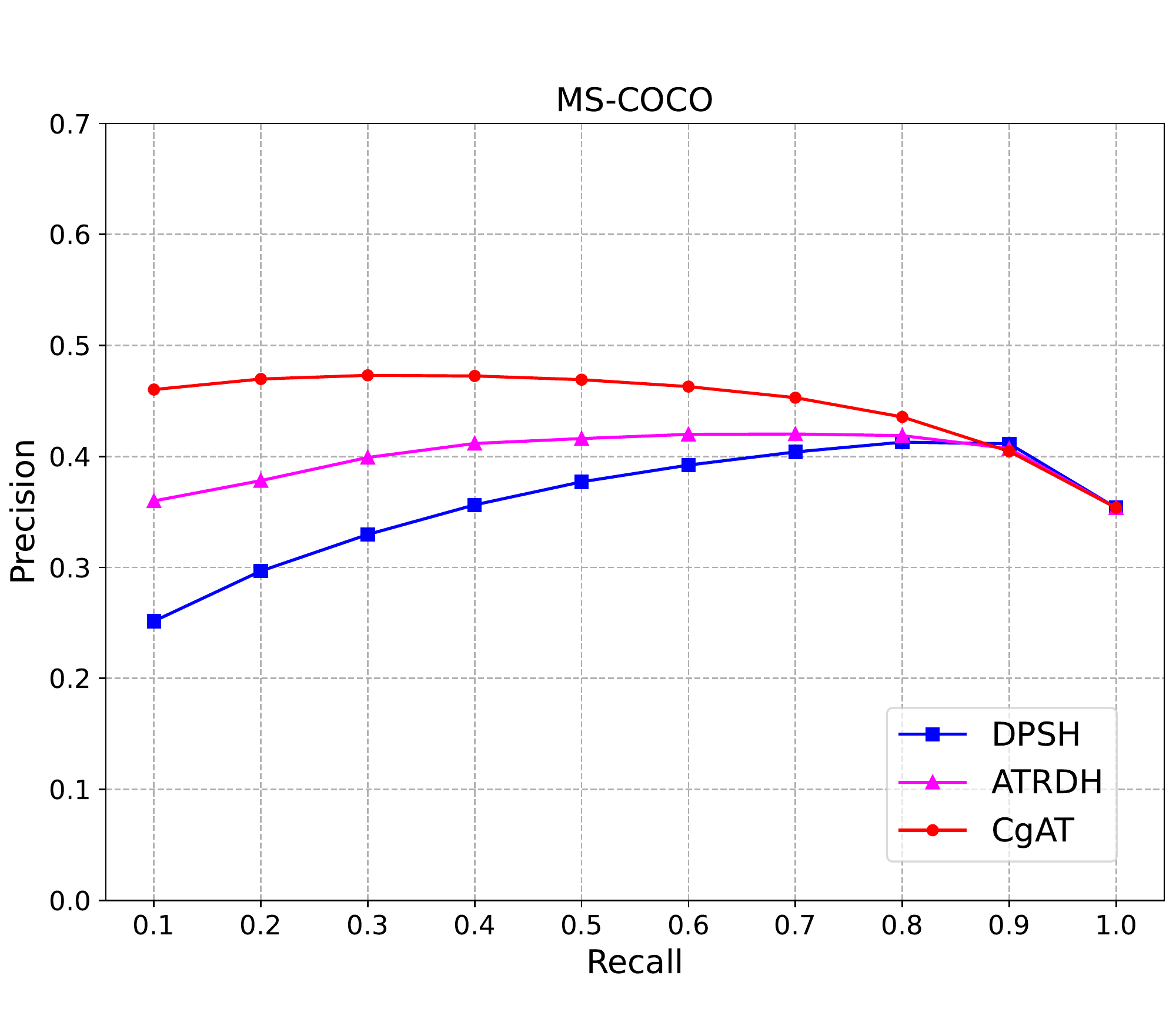}}
\vspace{-.5cm}
\caption{PR curves of defense methods under DHTA. These methods are based on deep hashing models with $K=32$.}
\label{fig:pr_dhta}
\end{figure*}

\begin{figure*}[ht]
\vspace{-0.6cm}
\centering
\subfigure[FLICKR-25K]{
    \includegraphics[width=0.31\textwidth]{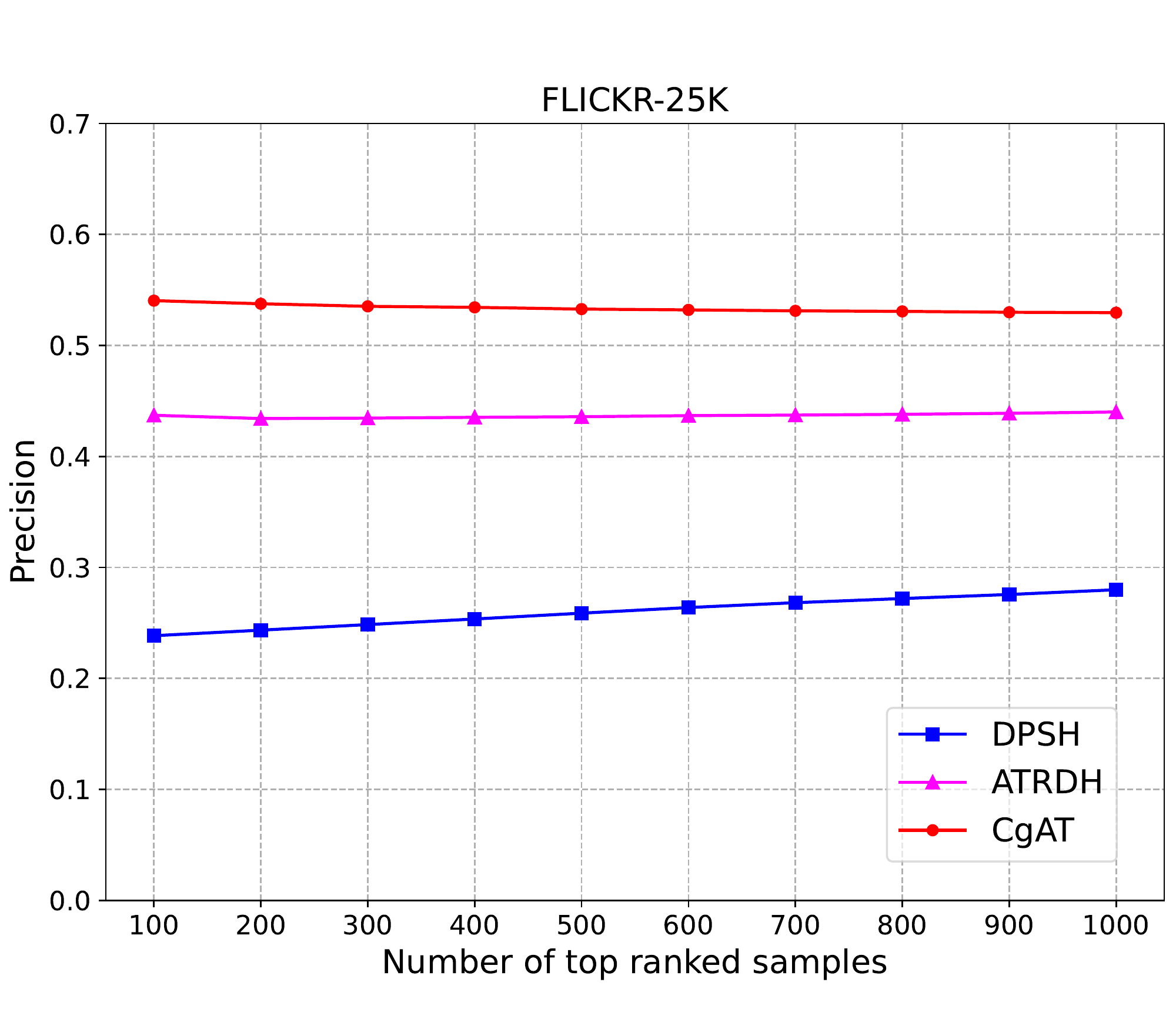}}
\subfigure[NUS-WIDE]{
    \includegraphics[width=0.31\textwidth]{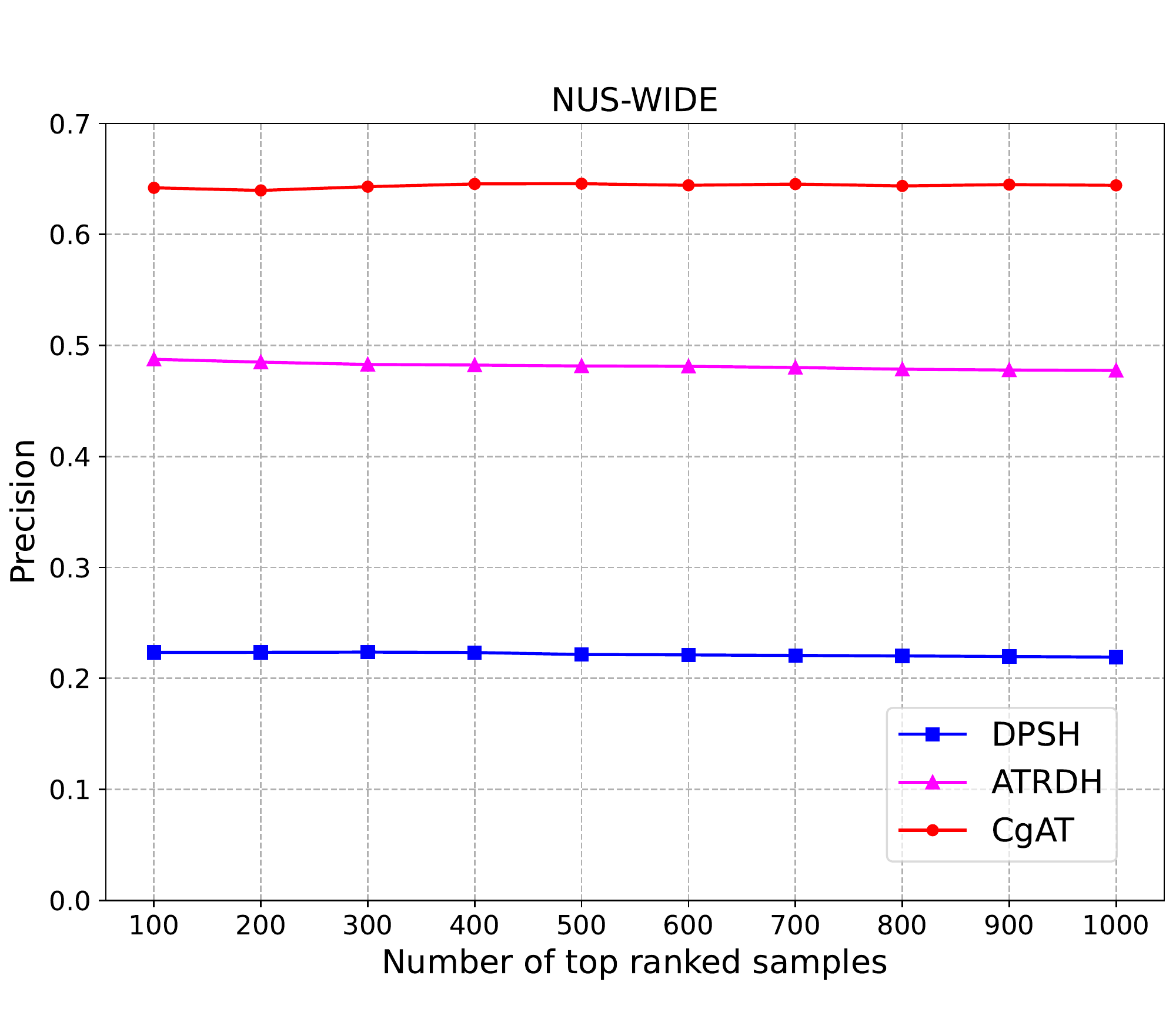}}
\subfigure[MS-COCO]{
    \includegraphics[width=0.31\textwidth]{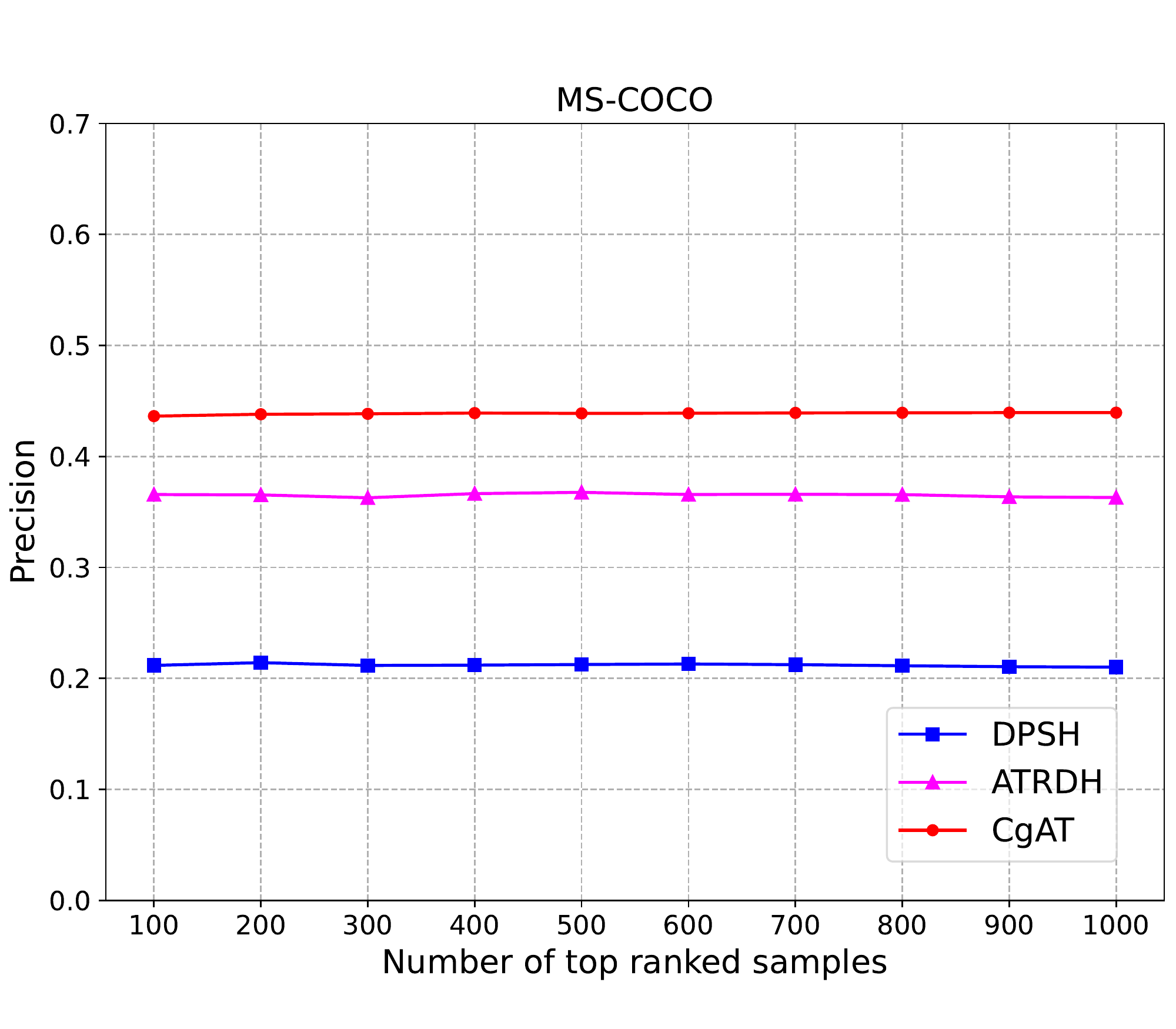}}
\vspace{-.5cm}
\caption{P@N curves for DHTA attacking adversarially-trained deep hashing models with 32 bits code length.}
\label{fig:topn_dhta}
\end{figure*}

\section{Experiments}
\subsection{Experimental Setup}
\textbf{Datasets.}
We adopt three popular datasets used in hashing-based retrieval to evaluate our defense method in extensive experiments: \textit{FLICKR-25K} \cite{huiskes2008mir}, \textit{NUS-WIDE} \cite{chua2009nus} and \textit{MS-COCO} \cite{lin2014microsoft}. The \textbf{FLICKR-25K} dataset comprises 25,000 Flickr images with 38 labels. We sample 1,000 images as the query set and the remaining regarded as the database, following \cite{wang2021targeted}. Moreover, we randomly select 5,000 instances from the database to train hashing models. The \textbf{NUS-WIDE} dataset contains 269,648 images annotated with 81 concepts. We sample a subset of the 21 most popular concepts, which consists of 195,834 images. 2,100 images are sampled from the subset as queries, while the rest images are regarded as the database. For the training set, we randomly select 10,500 images from the database \cite{wang2021prototype}. The \textbf{MS-COCO} dataset consists of 82,783 training samples and 40,504 validation samples, where each instance is annotated with at least one of the 80 categories. After combining the training and the validation set, we randomly pick 5,000 instances from them as queries and the rest as a database. For the training set, 10,000 images are randomly selected from the database.

\textbf{Protocols.}
To evaluate the defense performance of the proposed paradigm, we conduct experiments on the standard retrieval metrics, \textit{i.e.}, \textbf{Mean Average Precision} (\textbf{MAP}) \cite{yang2018adversarial}, \textbf{Precision-Recall} (\textbf{PR}) and \textbf{Precision@topN} (\textbf{P@N}) curves. Particularly, we calculate MAP values on the top 5000 retrieved results from the database \cite{yang2018adversarial, bai2020targeted}.

\textbf{Baselines.}
We adopt \textbf{DPSH} \cite{li2016feature} as the default hashing method for defense, which is one of the most typical and generic algorithms in deep hashing-based retrieval. Thus, $\mathcal{L}_{ori}$ in Eq. (\ref{eq:obj_cat}) for CgAT is the objective function of DPSH. In addition, we provide further experiments on other popular hashing methods to evaluate the generality of the proposed CgAT, including \textbf{HashNet} \cite{cao2017hashnet}, \textbf{DSDH} \cite{li2017deep}, \textbf{DCH} \cite{cao2018deep}, \textbf{CSQ} \cite{yuan2020central} and \textbf{DSDH-C} \cite{doan2022one} (refer to Table \ref{tab:univer}). VGG11 \cite{simonyan2014very} is selected as the default backbone network to implement hashing models on NUS-WIDE and MS-COCO, and AlexNet \cite{krizhevsky2012imagenet} is for FLICKR-25K. 
To evaluate the defense performance, we take multiple attack methods against the adversarially-trained models, including P2P \cite{bai2020targeted}, DHTA \cite{bai2020targeted}, ProS-GAN \cite{wang2021prototype}, THA \cite{wang2021targeted}, HAG \cite{yang2018adversarial} and SDHA \cite{lu2021smart}. For targeted attacks, we randomly select target labels that do not share the same category as the true labels. To be fair, other details of these methods are consistent with the original literature.

\begin{figure*}[t]
\centering
\vspace{-4mm}
\subfigure[FLICKR-25K]{ 
    \includegraphics[width=0.31\textwidth]{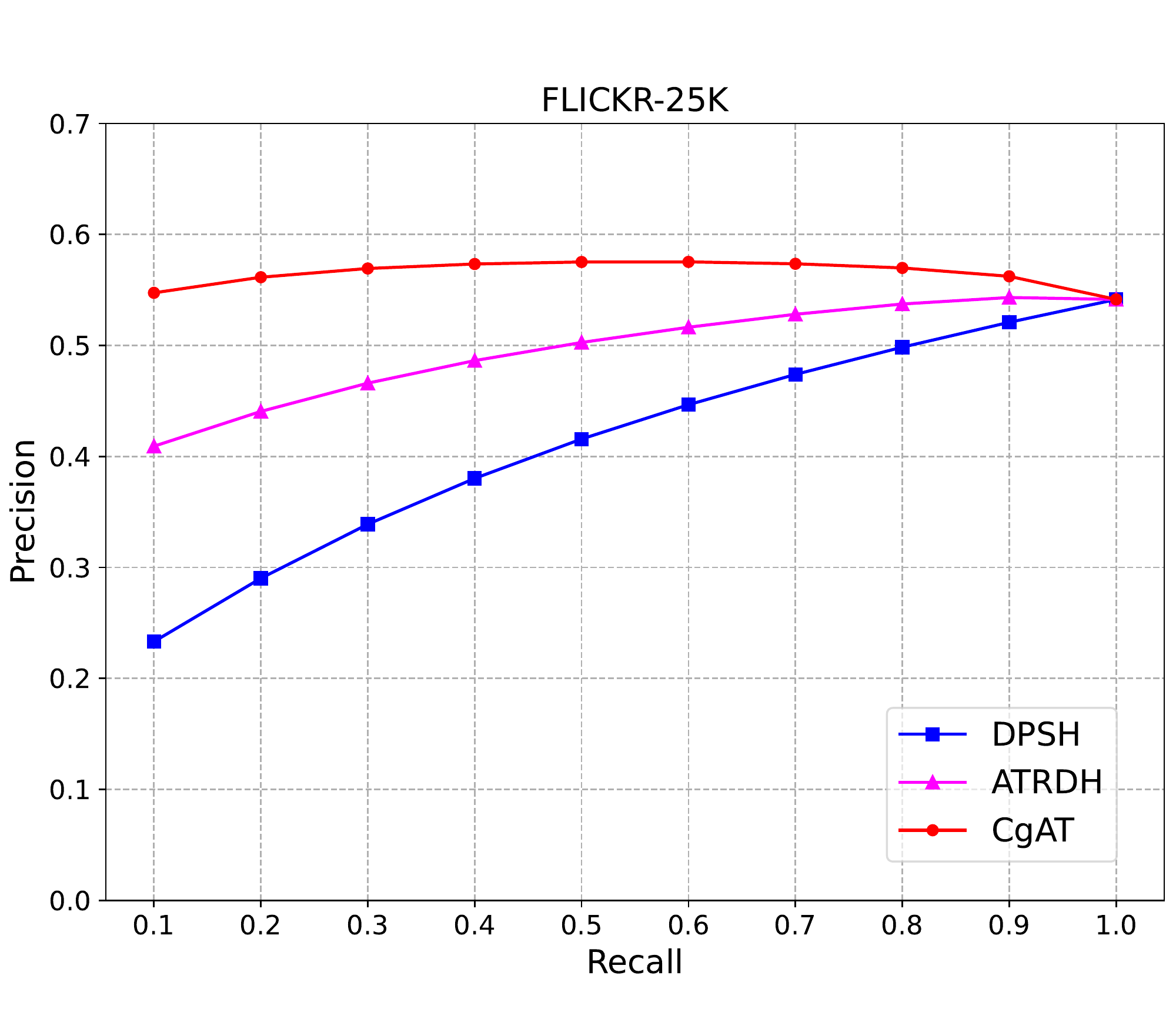}}
\subfigure[NUS-WIDE]{
    \includegraphics[width=0.31\textwidth]{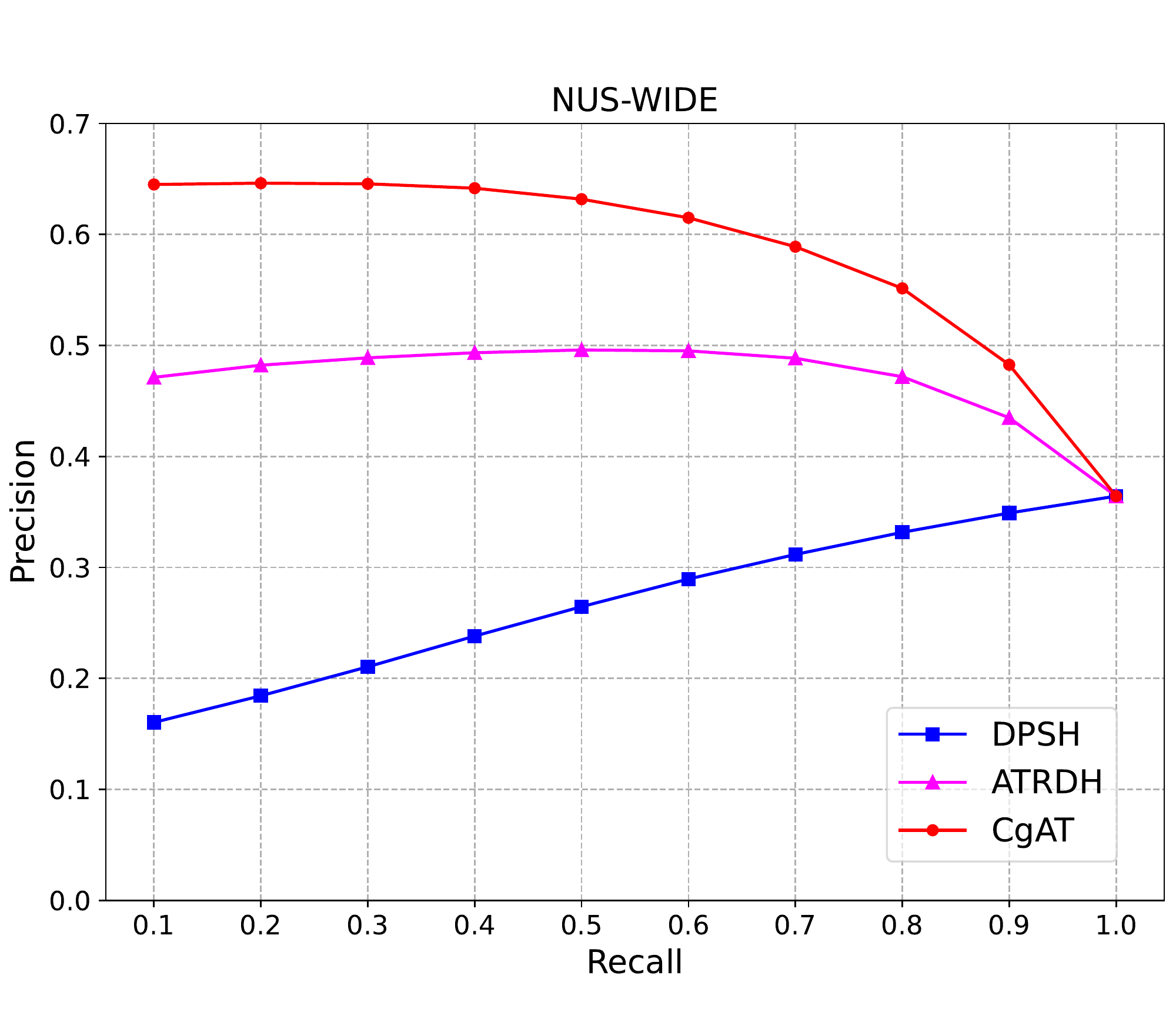}}  
\subfigure[MS-COCO]{
    \includegraphics[width=0.31\textwidth]{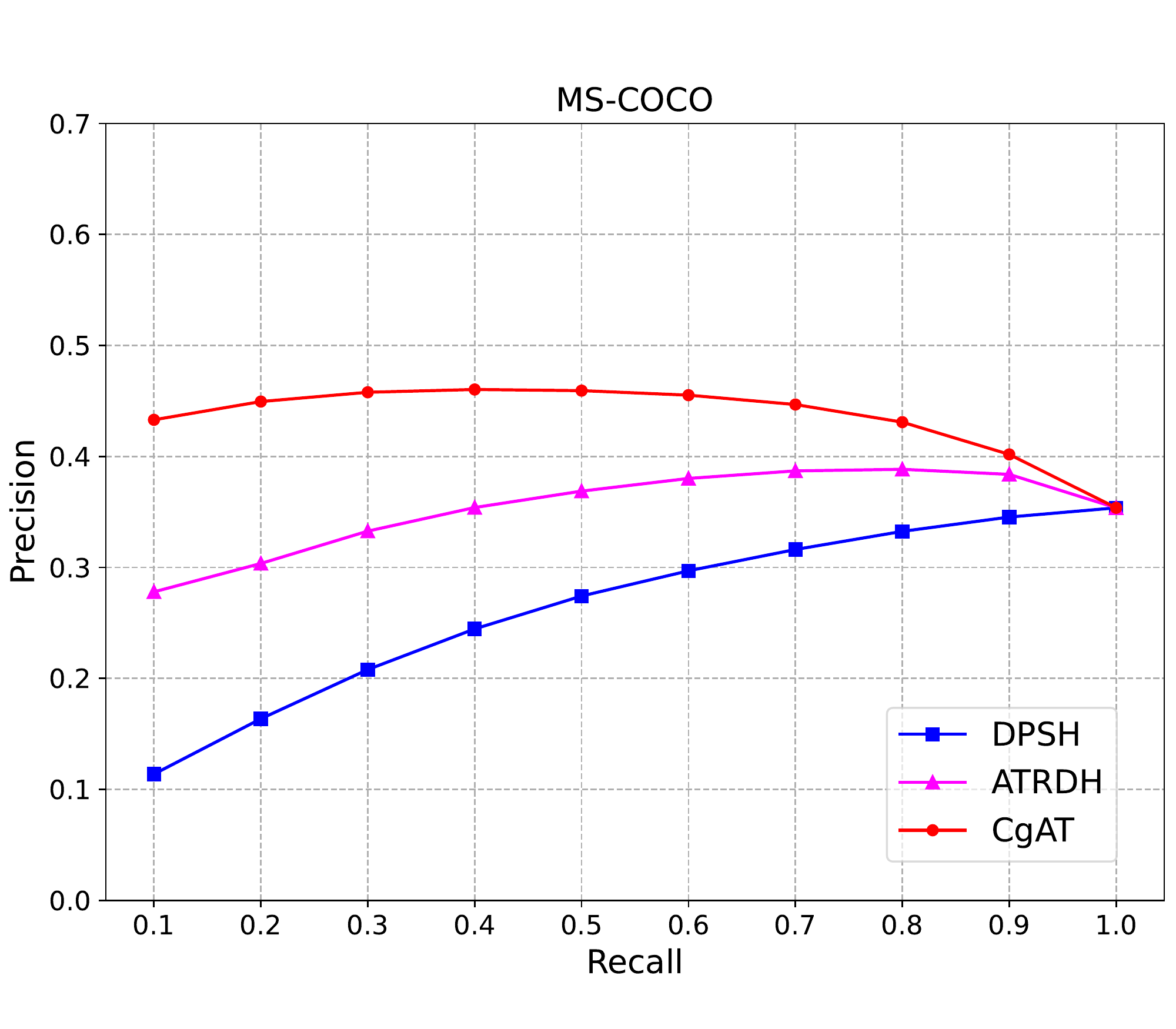}}
\vspace{-5mm}
\caption{PR curves of defense methods under SDHA.}
\label{fig:pr_sdha}
\end{figure*}

\begin{figure*}[t]
\centering
\vspace{-6mm}
\subfigure[FLICKR-25K]{
    \includegraphics[width=0.31\textwidth]{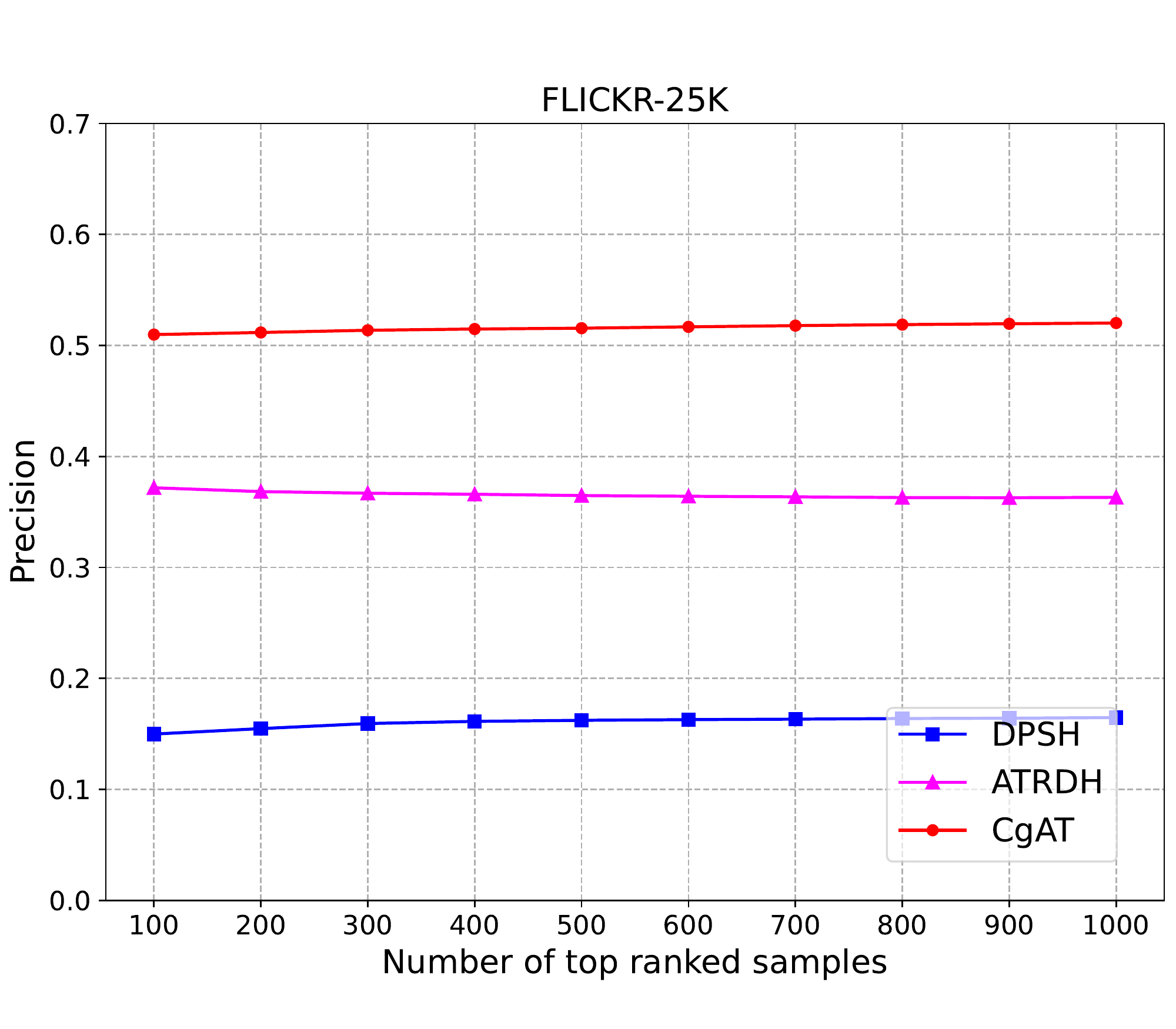}}
\subfigure[NUS-WIDE]{
    \includegraphics[width=0.31\textwidth]{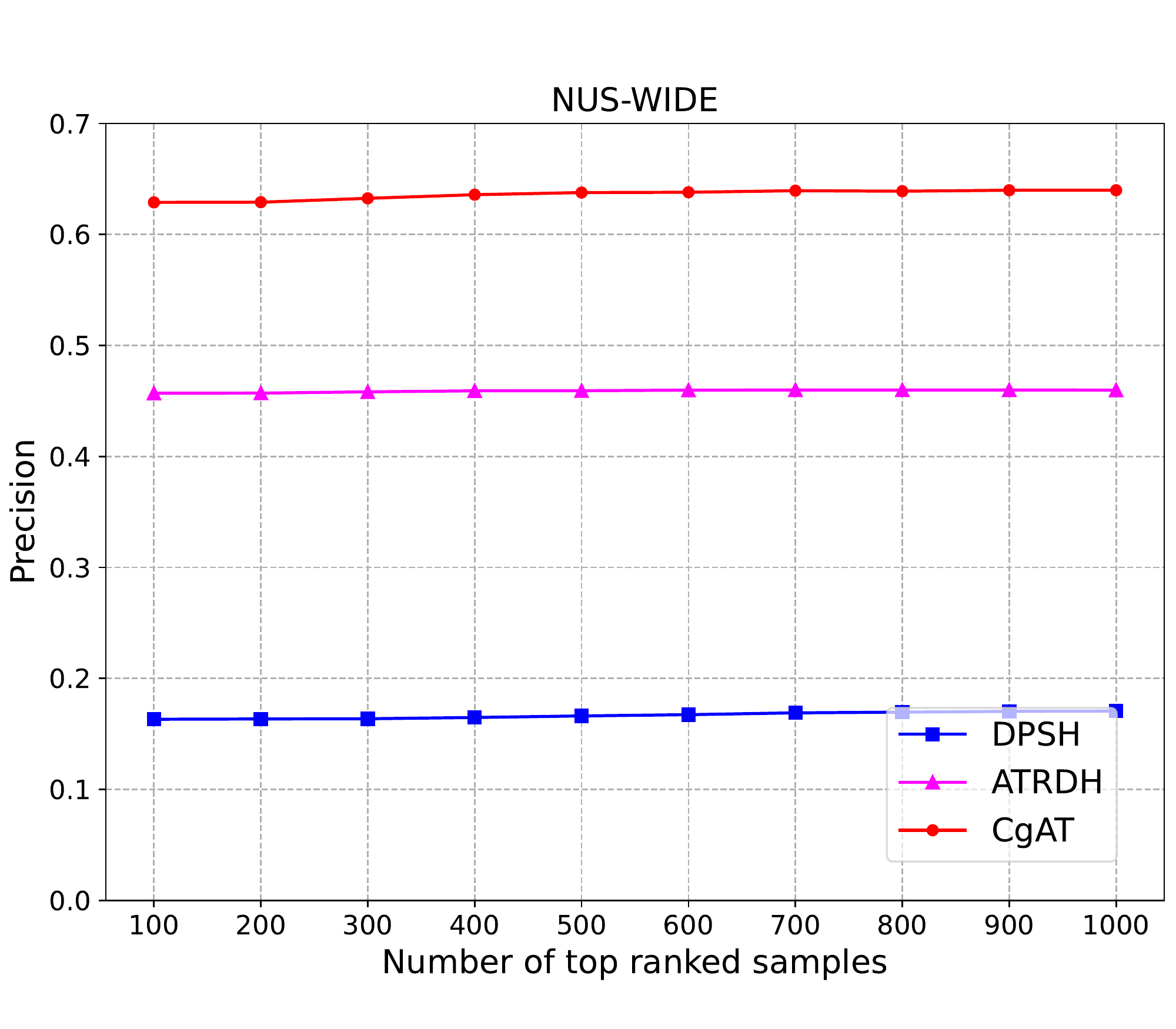}}  
\subfigure[MS-COCO]{
    \includegraphics[width=0.31\textwidth]{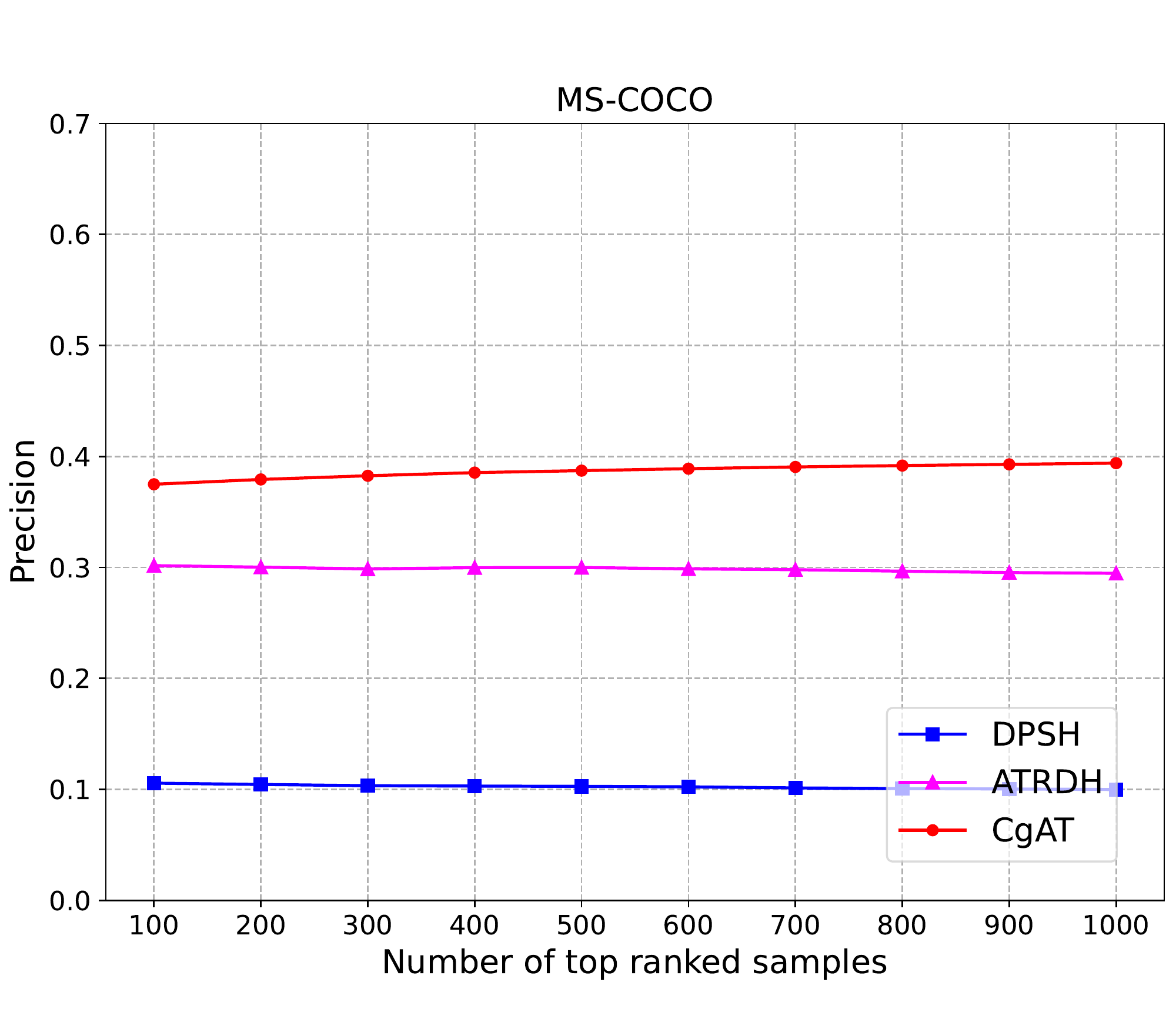}}
\vspace{-5mm}
\caption{P@N curves for SDHA attacking adversarially-trained deep hashing models with 32 bits code length.}
\label{fig:topn_sdha}
\end{figure*}

\textbf{Implementation Details.}
For target hashing models, we use SGD with initial learning rate $0.01$ and momentum $0.9$ as optimizer. We fix the batch size of images as $32$ and the weight decay as $5\times10^{-4}$. All images are resized to $224 \times 224$ and normalized in $[0,1]$ before feeding in hashing models. For our defense algorithm CgAT, we conduct 30 epochs for adversarial training. To generate adversarial examples in each training iteration, we set $\alpha$ and $T$ in PGD with $2/255$ and 7, respectively. The perturbation budget $\epsilon$ is set to $8/255$. The hyper-parameter $\lambda$ in Eq. \ref{eq:obj_cat} are set as $1$ by default. We conduct all the experimental evaluations using Pytorch 1.12 and run them on NVIDIA RTX 3090 GPUs.

\subsection{Defense Results}
We perform the proposed adversarial training algorithm on pre-trained deep hashing models to improve the adversarial robustness of deep hashing networks. After the adversarial training, we re-attack these models and the results in terms of MAP are reported in Table \ref{tab:flickr-25k}, \ref{tab:nus-wide} and \ref{tab:ms-coco}. Since DPSH \cite{li2016feature} is a typical deep hashing method and most of the mentioned methods are its variants, We develop CgAT and ATRDH based on it in these tables. "Clean" represents the MAP results tested by the original samples. As shown in the tables, the MAP values of ProS-GAN are similar to that of “Clean”, indicating that our defense method can completely resist the attack of ProS-GAN. By comparing our defense method with the original model (\textit{i.e.}, DPSH), we observe that the MAP values of different attack methods are much higher than no adversarial training, though the MAP values of the original (clean) samples decrease slightly. For example, under the targeted attack THA, CgAT brings an average increase of approximately 22.08\%, 40.19\%, and 18.30\% on FLICKR-25K, NUS-WIDE, and MS-COCO, respectively. For the non-targeted attack SDHA, our defense improves by an average of about 36.12\%, 44.19\%, and 30.86\% for FLICKR-25K, NUS-WIDE, and MS-COCO, respectively. These cases demonstrate that the proposed adversarial training strategy can effectively improve the robustness of deep hashing models against targeted and non-targeted attacks.

To further verify the effectiveness of our defense method, we compare our CgAT with ATRDH \cite{wang2021targeted} under the same experiment settings and the results are illustrated in Table \ref{tab:flickr-25k}, \ref{tab:nus-wide} and \ref{tab:ms-coco}. As shown in the tables, our defense method achieves a signiﬁcant performance boost in terms of the MAP under various attacks. For example, under the targeted attack THA, our defense method achieves an average improvement of over 5.02\%, 8.35\%, and 13.10\% for different bits on FLICKR-25K, NUS-WIDE, and MS-COCO, respectively, compared with ATRDH. Facing the non-targeted attack SDHA, our CgAT exceeds ATRDH by an average of 18.61\%, 12.35\%, and 11.56\% for FLICKR-25K, NUS-WIDE, and MS-COCO, respectively. The above phenomena show the proposed adversarial training algorithm can learn a more robust hashing model than ATRDH. 

In addition, we provide PR and P@N curves for DHTA and SDHA, as illustrated in Fig. \ref{fig:pr_dhta}, \ref{fig:topn_dhta}, \ref{fig:pr_sdha} and \ref{fig:topn_sdha}. The curves of CgAT consistently lie above other methods, which further indicates the superior defense performance of our proposed method. 


\begin{table*}[t]
\caption{MAP (\%) under different attacks with $\epsilon=8/255$. The lower the MAP, the stronger the attack performance. For our attack, we set $\alpha$ and $T$ of PGD \cite{madry2017towards} to $1/255$ and $100$, respectively.}
\vspace{-0.3cm}
\centering
\resizebox{0.95\textwidth}{!}
{
\begin{tabular}{l|ccc|ccc|ccc}
\toprule
\multirow{2}{*}{Attack Method} & \multicolumn{3}{c|}{FLICKR-25K} & \multicolumn{3}{c|}{NUS-WIDE} & \multicolumn{3}{c}{MS-COCO} \\ \cline{2-10}
&DPSH \cite{li2016feature} &ATRDH \cite{wang2021targeted} &CgAT (ours) &DPSH \cite{li2016feature} &ATRDH \cite{wang2021targeted} &CgAT (ours) &DPSH \cite{li2016feature} &ATRDH \cite{wang2021targeted} &CgAT (ours)   \\ \hline
Clean &84.00 &76.65 &77.12 &82.20 &70.38 &72.05 &60.50 &48.04 &54.13 \\ \hline
P2P \cite{bai2020targeted}   &40.26 &52.04 &58.07 &29.50 &48.31 &64.81 &25.99 &36.65 &44.51 \\
DHTA \cite{bai2020targeted}  &32.78 &47.25 &53.68 &23.38 &47.96 &64.56 &21.34 &35.74 &44.15 \\
ProS-GAN \cite{wang2021prototype} &71.81 &77.14 &77.51 &29.84 &69.64 &71.07 &44.06 &48.00 &53.97 \\
THA \cite{wang2021targeted}  &36.17 &50.67 &57.09 &21.79 &51.14 &63.02 &43.88 &33.60 &51.08 \\
HAG \cite{yang2018adversarial}   &26.72 &38.24 &49.91 &17.18 &47.17 &64.62 &9.22 &29.18 &47.40 \\
SDHA \cite{lu2021smart}  &19.34 &39.18 &53.42 &17.82 &46.14 &64.22 &10.07 &29.47 &40.37 \\ \hline
Ours  &\textbf{17.04} &\textbf{27.96} &\textbf{37.89} &\textbf{9.93} &\textbf{30.83} &\textbf{54.36} &\textbf{8.87} &\textbf{22.57} &\textbf{34.15} \\ \bottomrule
\end{tabular}
}
\label{tab:attack}
\end{table*}

\subsection{Discussion}
\begin{figure}[ht]
    \vspace{-2mm}
	\begin{center}
		\includegraphics[width=0.8\linewidth]{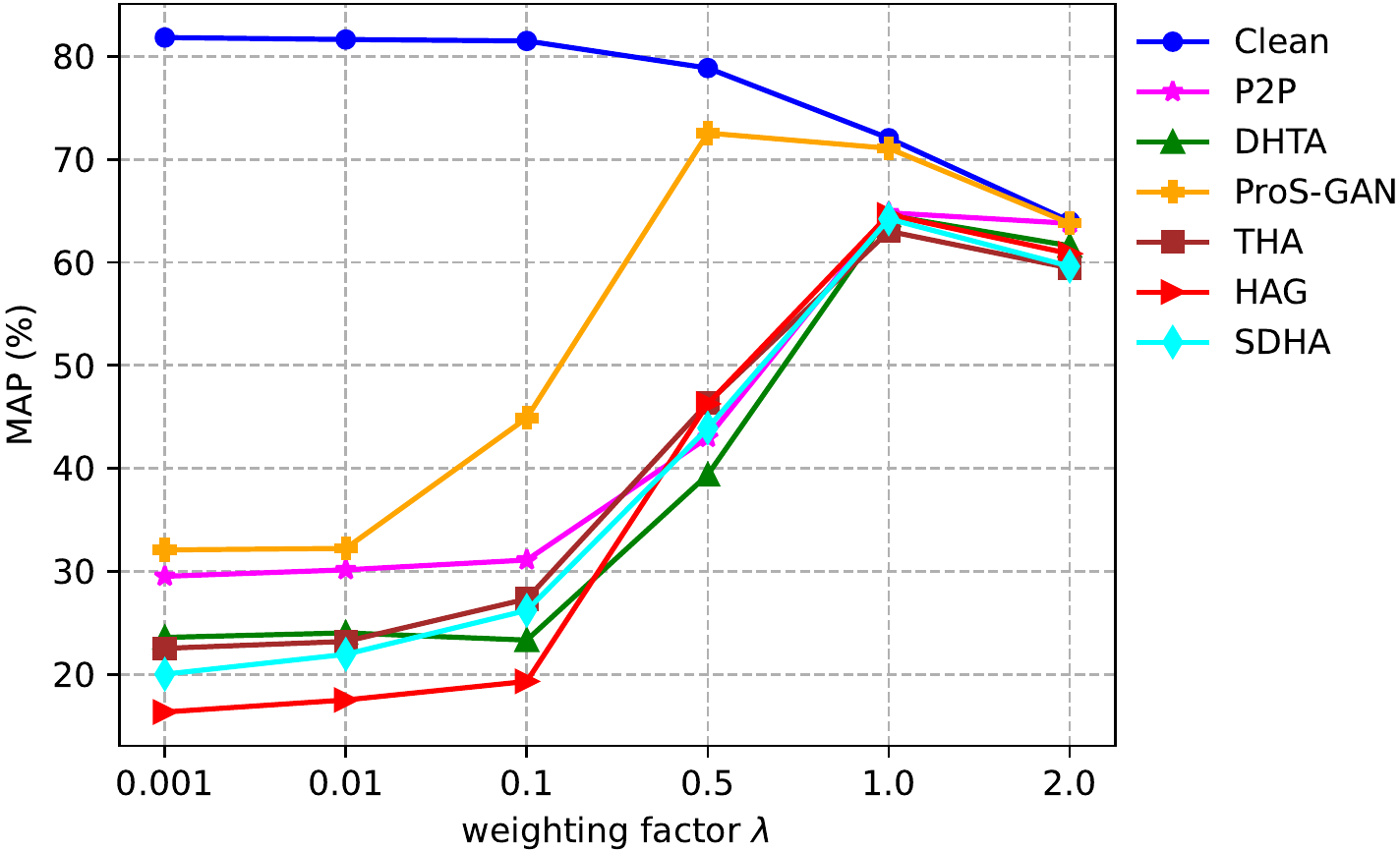}
	\end{center}
	\vspace{-.3cm}
	\caption{MAP (\%) on NUS-WIDE for our adversarial training with different $\lambda$.}
	\label{fig:lambda}
	\vspace{-3mm}
\end{figure}

\textbf{Effect of $\lambda$}.
The hyper-parameter $\lambda$ controls the quality of the adversarial training. To explore the effect of different weighting factors on defense performance, we make comparison experiments with 32-bits hashing model, as illustrated in Fig. \ref{fig:lambda}. As shown in Fig. \ref{fig:lambda}, when $\lambda$ increases, the defense performance increases, but the MAP values of original samples drop, which indicates there is a trade-off between robustness and precision. When $\lambda$ is too large (\textit{e.g.}, $\lambda=2$), the defense performance and the precision of clean samples are reduced simultaneously. Therefore, robustness and precision affect each other in deep hashing.

\textbf{Attack performance of our adversarial examples}.
We argue that the adversarial examples generated by our method are more powerful than others, which benefits adversarial training in enhancing the defense performance of deep hashing networks. To verify this point, we explore the attack performance of the proposed way to construct adversarial samples. The attack results on three datasets are shown in Table \ref{tab:attack}. From the results, our attack is stronger than all other methods. In contrast to the cases on original hashing models without defense, adversarial examples generated by our method have obvious advantages than the state-of-the-art HAG and SDHA in attacking defense models. Besides, we note that the proposed CgAT significantly outperforms other methods under our attack, which further confirms the effectiveness of CgAT.

\subsection{Universality on different hashing methods}
To validate that the proposed adversarial training paradigm is generic to most popular hashing models, We apply CgAT to different hashing methods (\textit{i.e.}, replace $\mathcal{L}_{ori}$ in Eq. (\ref{eq:obj_cat}) with the objective function of the corresponding hashing method). The results are summarized in Table \ref{tab:univer}. Moreover, even with different hashing methods, our defense method can still effectively mitigate the impact of adversarial attacks. Furthermore, when testing with various attacks, the results of CgAT are higher than ATRDH in almost cases, which shows that hashing models trained by our CgAT are more robust than ATRDH. Hence, the above phenomena confirm the universality of the proposed defense method.

\begin{table}[ht]
\caption{MAP for different hashing models on NUS-WIDE. CgAT and ATRDH \cite{wang2021targeted} are built on original hashing models, \textit{e.g.}, we replace $\mathcal{L}_{ori}$ in Eq. (\ref{eq:obj_cat}) with the objective function of HashNet \cite{cao2017hashnet}.}
\vspace{-0.3cm}
\centering
\resizebox{\columnwidth}{!}{
\begin{tabular}{l|c|cccccc}
\toprule
Method     &Clean &P2P &DHTA &ProS-GAN &THA &HAG &SDHA \\ \hline
HashNet \cite{cao2017hashnet}   &79.57 &31.40 &25.12 &28.62 &19.65  &14.56 &13.99  \\
-ATRDH     &68.48 &56.43 &54.46 &69.33 &57.00  &48.25 &48.67  \\
-CgAT     &70.26 &\textbf{64.64} &\textbf{64.02} &\textbf{70.28} &\textbf{61.23} &\textbf{57.79} &\textbf{60.24}   \\ \hline
DSDH \cite{li2017deep}  &81.90 &29.63 &22.30 &30.05 &23.57 &16.92 &18.96  \\
-ATRDH     &68.77 &54.35 &53.21 &70.03 &55.80 &50.96 &50.94   \\
-CgAT     &73.52 &\textbf{62.29} &\textbf{60.82} &\textbf{72.30} &\textbf{61.23} &\textbf{62.12}  &\textbf{63.01}   \\ \hline
DCH \cite{cao2018deep}  &77.43 &37.04  &31.68 &62.87 &27.18 &25.28 &27.06   \\
-ATRDH     &70.50 &41.10 &37.96 &58.21 &33.58 &29.47 &47.42  \\
-CgAT     &73.10 &\textbf{59.95} &\textbf{58.56} &\textbf{73.26} &\textbf{56.94} &\textbf{65.71} &\textbf{60.17}   \\ \hline
CSQ \cite{yuan2020central}  &80.08 &32.89 &27.35 &31.01 &27.09 &19.91 &13.88   \\
-ATRDH     &75.52 &35.86 &31.78 &\textbf{73.72} &34.72 &18.34 &21.24   \\
-CgAT     &75.43 &\textbf{50.08} &\textbf{48.79} &{65.77} &\textbf{46.76} &\textbf{46.30} &\textbf{52.76}  \\ \hline
DSDH-C \cite{doan2022one}  &83.70 &29.54 &22.12 &32.39 &25.50 &15.39 &12.59  \\
-ATRDH     &72.63 &48.47 &45.47 &73.40 &46.48 &40.71 &43.73   \\
-CgAT      &73.06 &\textbf{67.43} &\textbf{66.66} &\textbf{73.64} &\textbf{64.45} &\textbf{68.29} &\textbf{63.60}  \\
\bottomrule
\end{tabular}
}
\label{tab:univer}
\end{table}

\section{Conclusion}
In this paper, we proposed the adversarial training algorithm (\textit{i.e.}, CgAT) for deep hashing-based retrieval. Specifically, we first provided the continuous hash center method to obtain the center code as the optimal representative of the image semantics for helping construct the adversarial learning framework. Moreover, we took the center code as 'label' to guide the generation of strong adversarial samples, where the similarity between the center code and the hash code of the adversarial example was minimized. Furthermore, the adversarial training maximized the similarity of the adversarial sample to the center code to improve the adversarial robustness of deep hashing networks for defense. Experiments showed that our method performed state-of-the-art results in defense of deep hashing-based retrieval.

\begin{acks}
This work was supported in part by the Hong Kong Innovation and Technology Fund under Project ITS/030/21 and in part by the research grant from Shenzhen Municipal Central Government Guides Local Science and Technology Development Special Funded Projects under Grant 2021Szvup139.
\end{acks}

\bibliographystyle{ACM-Reference-Format}
\bibliography{sample-base}

\clearpage
\appendix

\section{Proof of CHCM}
\label{sec:proof}
Center code $\boldsymbol{b}^{\ast}$ which satisfies Eq. (\ref{eq:obj_center_code}) can be calculated by the Continuous Hash Center Method (CHCM), \textit{i.e.}, 
\begin{equation*}
    \begin{aligned}
    \boldsymbol{b}^{\ast} &= \arg\min_{\boldsymbol{b}\in\{-1,+1\}^K} \sum_{i}^{N_{\rm{p}}} w_i {D}_{\rm{H}}(\boldsymbol{b}, \boldsymbol{b}_i^{(\rm{p})}) - \sum_{j}^{N_{\rm{n}}} w_j {D}_{\rm{H}}(\boldsymbol{b}, \boldsymbol{b}_j^{(\rm{n})}) \\
    &=\operatorname{sign}\left(\sum_{i}^{N_{\rm{p}}}w_{i}\boldsymbol{b}_i^{(\rm{p})} - \sum_{j}^{N_{\rm{n}}}w_{j}\boldsymbol{b}_j^{(\rm{n})} \right).
    \end{aligned}
\end{equation*}

\begin{proof}
We define the following function:
\begin{equation*}
    \begin{aligned}
    \psi(\boldsymbol{b})=\sum_{i} w_i {D}_{\rm{H}}(\boldsymbol{b}, \boldsymbol{b}_i^{(\rm{p})}) - \sum_{j} w_j {D}_{\rm{H}}(\boldsymbol{b}, \boldsymbol{b}_j^{(\rm{n})})
    \end{aligned}.
\end{equation*}
As the center code $\boldsymbol{b}^{\ast}$ need to be the optimal solution of the minimizing objective, the above theorem is equivalent to prove the following inequality:
\begin{equation*}
    \begin{aligned}
    \psi(\boldsymbol{b})\geq \psi(\boldsymbol{b}^{\ast}),
    \quad \forall~\boldsymbol{b}\in\{-1,+1\}^K
    \end{aligned}.
\end{equation*}
Let $\boldsymbol{b}=\{b_1,b_2,...,b_K\}$, then we have
\begin{equation*}
    \begin{aligned}
    &\psi(\boldsymbol{b})
    =\sum_{i} w_i \frac{1}{2}(K-\boldsymbol{b}^\top \boldsymbol{b}_i^{(\rm{p})}) - \sum_{j} w_j \frac{1}{2}(K-\boldsymbol{b}^\top \boldsymbol{b}_j^{(\rm{n})})\\
    =&-\frac{1}{2}\sum_{i} w_i\boldsymbol{b}^\top \boldsymbol{b}_i^{(\rm{p})}+\frac{1}{2}\sum_{j} w_j\boldsymbol{b}^\top \boldsymbol{b}_j^{(\rm{n})} + \xi \\
    =&-\frac{1}{2}\sum_{i}w_i\sum_{k=1}^K{b}_k{b}_{ik}^{(\rm{p})}+\frac{1}{2}\sum_{j} w_j\sum_{k=1}^K{b}_k{b}_{jk}^{(\rm{p})}+\xi \\
    =&-\frac{1}{2}\sum_{k=1}^K b_k(\sum_{i}w_ib_{ik}^{(\rm{p})}-\sum_{j}w_jb_{jk}^{(\rm{n})})+\xi, \\
    \end{aligned}
\end{equation*}
where $\xi$ is a constant.
Similarly, 
\begin{equation*}
    \begin{aligned}
    \psi(\boldsymbol{b}^{\ast})=-\frac{1}{2}\sum_{k=1}^K b^{\ast}_{k}(\sum_{i}w_ib_{ik}^{(\rm{p})}-\sum_{j}w_jb_{jk}^{(\rm{n})})+\xi.
    \end{aligned}
\end{equation*}
Let $\Omega=\sum_{i}w_ib_{ik}^{(\rm{p})}-\sum_{j}w_jb_{jk}^{(\rm{n})}$, then
\begin{equation*}
\begin{aligned}
\psi(\boldsymbol{b})=-\frac{1}{2}\sum_{k=1}^K b_{k}\Omega+\xi, \quad
\psi(\boldsymbol{b}^{\ast})=-\frac{1}{2}\sum_{k=1}^K b^{\ast}_{k}\Omega+\xi.
\end{aligned}
\end{equation*}
Due to the nature of absolute value, we have
\begin{equation*}
    \begin{aligned}
    &\psi(\boldsymbol{b})
    =-\frac{1}{2}\sum_{k=1}^K b_k\Omega+\xi \\
    &\geq-\frac{1}{2}\sum_{k=1}^K \left|b_k\Omega\right|+\xi =-\frac{1}{2}\sum_{k=1}^K \left|\Omega\right|+\xi \\
    &=-\frac{1}{2}\sum_{k=1}^K \operatorname{sign}(\Omega)\Omega+\xi \\
    &=-\frac{1}{2}\sum_{k=1}^K b^{\ast}_{k}\Omega+\xi \\ &=\psi(\boldsymbol{b}^{\ast}).
    \end{aligned}
\end{equation*}
That is, $\psi(\boldsymbol{b})\geq\psi(\boldsymbol{b}^{\ast})$. Thus, Theorem \ref{theo:chcm} is proved.
\end{proof}


\end{document}